\documentclass[10pt,twocolumn,letterpaper]{article}

\usepackage{iccv}
\usepackage{times}
\usepackage{epsfig}
\usepackage{graphicx}
\usepackage{amsmath}
\usepackage{amssymb}
\usepackage{booktabs}
\usepackage{multirow}
\usepackage{pstricks}
\usepackage[binary-units=true]{siunitx}
\usepackage{capt-of}
\usepackage{etoolbox}
\usepackage{tabularx} 
\usepackage{xcolor}

\makeatletter\let\captiontemp\@makecaption\makeatother
\usepackage[font=footnotesize,labelformat=simple]{subcaption}
\makeatletter\let\@makecaption\captiontemp\makeatother

\usepackage{amsmath,amsfonts,bm}

\def\eqref#1{equation~\ref{#1}}

\def\1{\bm{1}}

\newcommand{\transpose}{^\mathsf{T}}

\def\rva{{\mathbf{a}}}
\def\rvb{{\mathbf{b}}}

\def\rvp{{\mathbf{p}}}

\def\rvx{{\mathbf{x}}}
\def\rvy{{\mathbf{y}}}

\def\rmW{{\mathbf{W}}}

\DeclareMathAlphabet{\mathsfit}{\encodingdefault}{\sfdefault}{m}{sl}
\SetMathAlphabet{\mathsfit}{bold}{\encodingdefault}{\sfdefault}{bx}{n}

\newcommand{\R}{\mathbb{R}}

\definecolor{forestgreen}{rgb}{0.0, 0.27, 0.13}
\definecolor{frenchblue}{rgb}{0.0, 0.45, 0.73}
\definecolor{electricviolet}{rgb}{0.56, 0.0, 1.0}
\definecolor{electriccrimson}{rgb}{1.0, 0.0, 0.25}
\definecolor{hanpurple}{rgb}{0.32, 0.09, 0.98}

\usepackage[breaklinks=true,bookmarks=false]{hyperref}

\iccvfinalcopy 

\def\iccvPaperID{1797} 

\ificcvfinal\fi

\makeatletter
\apptocmd\@maketitle{{\myfigure{}\vspace{12pt}\par}}{}{}
\makeatother

\begin{document}

\title{Learning to Estimate Hidden Motions with Global Motion Aggregation}

\author{Shihao Jiang\textsuperscript{1,2} \qquad  
Dylan Campbell\textsuperscript{3} \qquad
Yao Lu\textsuperscript{1,2}  \qquad
Hongdong Li\textsuperscript{1,2}  \qquad 
Richard Hartley\textsuperscript{1,2}\\
\\
\textsuperscript{1}Australian National University \qquad 
\textsuperscript{2}ACRV \qquad 
\textsuperscript{3}University of Oxford}

\newcommand\myfigure{%
\centering
\includegraphics[width=\textwidth]{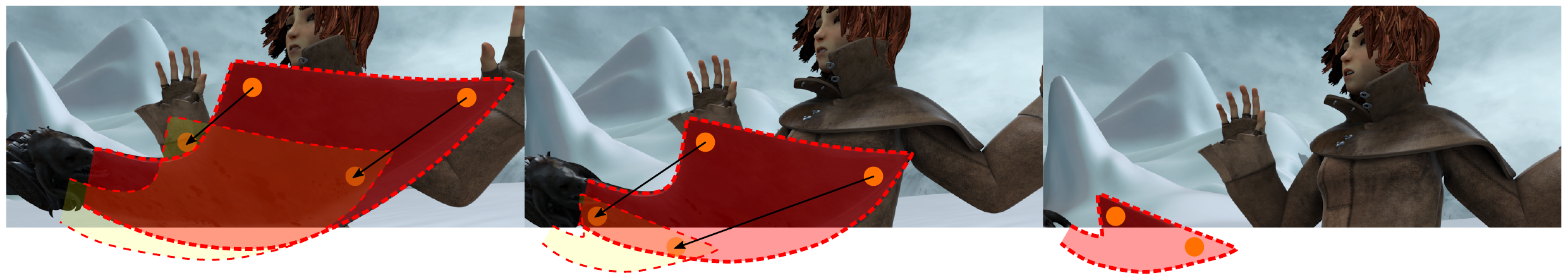}
\captionof{figure}{%
\textbf{Global motion aggregation helps resolve ambiguities caused by 
occlusions.}
Occlusions---a term we extend to include any parts of a scene that disappear in the next frame---cause large ambiguities in the optical flow estimation problem that cannot be resolved by local approaches. Based on the assumption that points on an object have homogeneous motions, which often holds approximately, we propose to globally aggregate motion features of pixels that are likely to belong to the same object.
In this example, most pixels on the blade move out-of-frame from frame 2 to frame 3. When only these two frames are provided, global aggregation allows motion information to be passed from non-occluded pixels to occluded pixels, which helps resolve ambiguities caused by occlusions.
}
\label{fig:demo}
}

\maketitle
\ificcvfinal\thispagestyle{empty}\fi

\begin{abstract}
\vspace{-7pt}
Occlusions pose a significant challenge to optical flow 
algorithms that rely on local evidences. We consider an occluded point to be one
that is imaged in the reference frame but not in the next, a slight overloading of
the standard definition since it also includes points that move out-of-frame. 
Estimating the motion of these points is extremely difficult, particularly
in the two-frame setting.
Previous work relies on CNNs to learn occlusions, without much success, or
requires multiple frames to reason about occlusions using temporal smoothness.
In this paper, we argue that the occlusion problem can be better 
solved in the two-frame case by modelling image self-similarities. 
We introduce a global motion aggregation module, a transformer-based
approach to find long-range dependencies between pixels in the first image,
and perform global aggregation on the corresponding motion features. 
We demonstrate that the optical flow estimates in the occluded regions can be
significantly improved without damaging the performance in non-occluded regions.
This approach obtains new state-of-the-art results on the challenging Sintel dataset,
improving the average end-point error by 13.6\% on Sintel Final and 13.7\% on Sintel Clean.
At the time of submission, our method ranks first on these benchmarks among all
published and unpublished approaches. 
Code is available at \url{https://github.com/zacjiang/GMA}.
\end{abstract}

\section{Introduction}
\label{Sec:intro}

How can we estimate the 2D motion of a point we only see once? This is the problem
faced by optical flow algorithms for points that become occluded between frames.
Estimating the optical flow, that is, the apparent motion of pixels in an image
as the camera and scene move, is a classic problem in computer vision studied since
the seminal work of Horn and Schunck \cite{horn}.
There are many factors that make optical flow prediction a hard problem, including
large motions, motion and defocus blur, and featureless regions. Among these challenges,
occlusion is one of the most difficult and under-explored. In this paper, we
propose an approach that specifically targets the occlusion problem in the case of two-frame
optical flow prediction. 

We first define what we mean by occlusion in the context of optical flow estimation.
In this paper, an occluded point is defined as a 3D point that is imaged in the reference
frame but is not visible in the matching frame. This definition incorporates several
different scenarios, such as the query point moving out-of-frame or behind another
object (or itself), or another object moving in front of the query point, in the
active sense. One particular case of occlusion is shown in Figure~\ref{fig:demo},
where part of the blade moves out-of-frame.

\begin{figure*}[t!]
     \centering
     \begin{subfigure}[b]{0.141\textwidth}
         \centering
         \caption*{Reference Frame}
         \includegraphics[width=\textwidth, trim=0 0 250pt 0, clip]{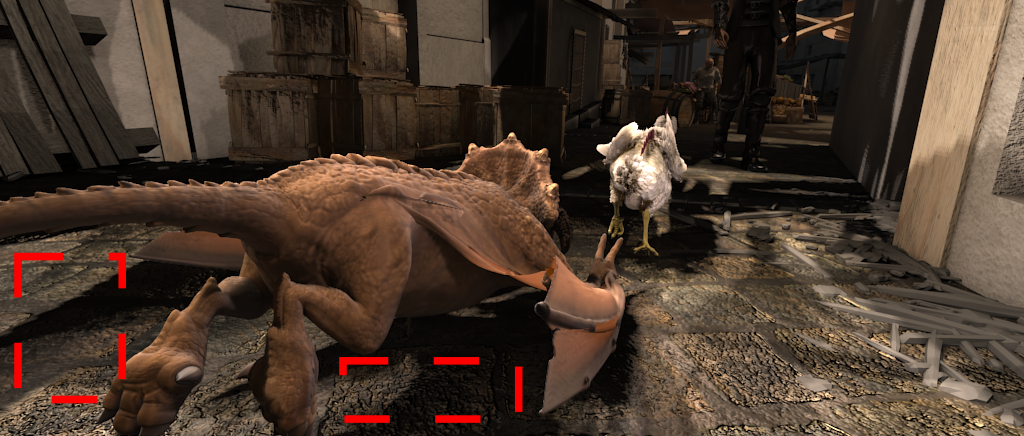}
     \end{subfigure}%
     \hfill
     \begin{subfigure}[b]{0.141\textwidth}
         \centering
         \caption*{Matching Frame}
         \includegraphics[width=\textwidth, trim=0 0 250pt 0, clip]{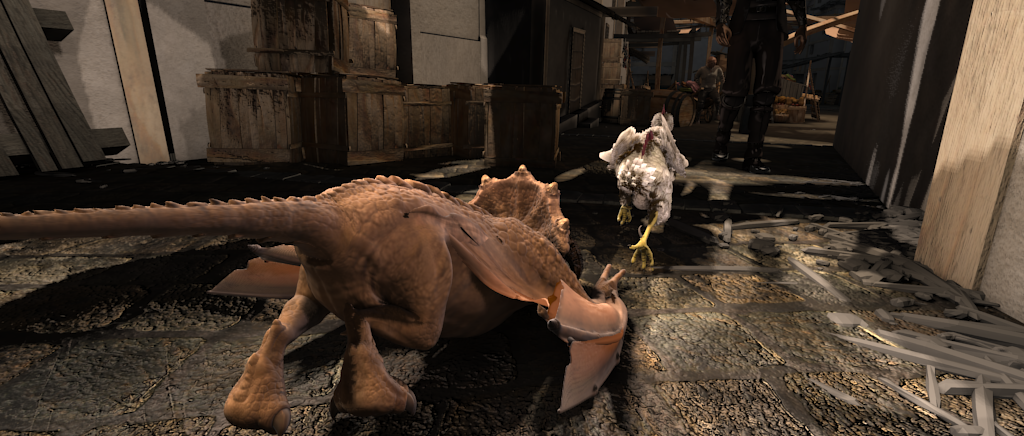}
     \end{subfigure}%
     \hfill
     \begin{subfigure}[b]{0.141\textwidth}
         \centering
         \caption*{Flow (RAFT \cite{raft})}
         \includegraphics[width=\textwidth, trim=0 0 250pt 0, clip]{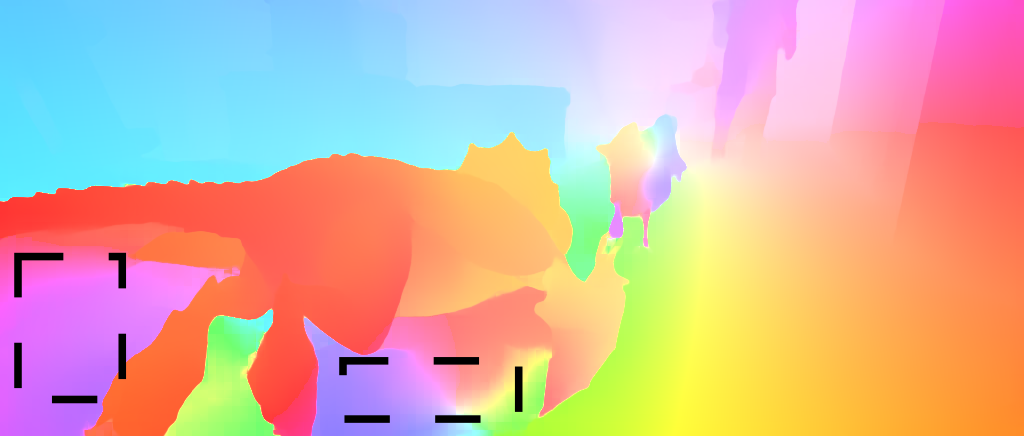}
     \end{subfigure}%
     \hfill
     \begin{subfigure}[b]{0.141\textwidth}
         \centering
         \caption*{Flow (Ours)}
         \includegraphics[width=\textwidth, trim=0 0 250pt 0, clip]{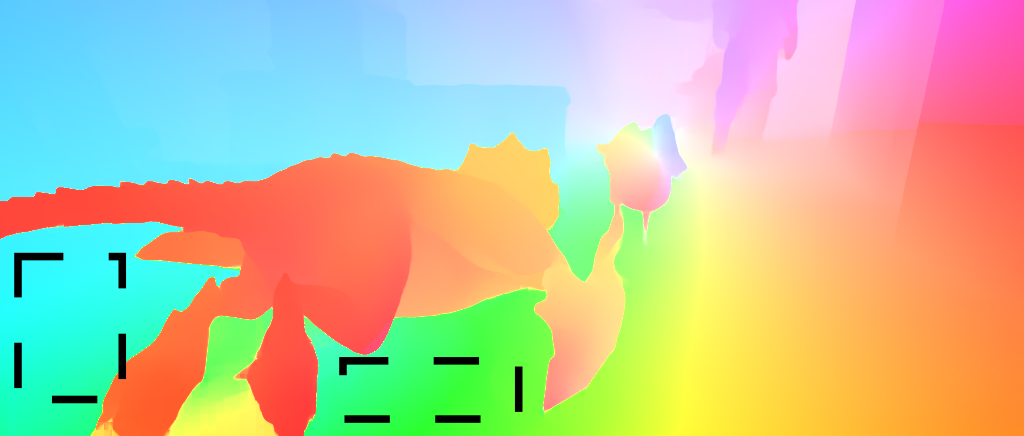}
     \end{subfigure}%
     \hfill
     \begin{subfigure}[b]{0.141\textwidth}
         \centering
         \caption*{Flow (Ground Truth)}
         \includegraphics[width=\textwidth, trim=0 0 250pt 0, clip]{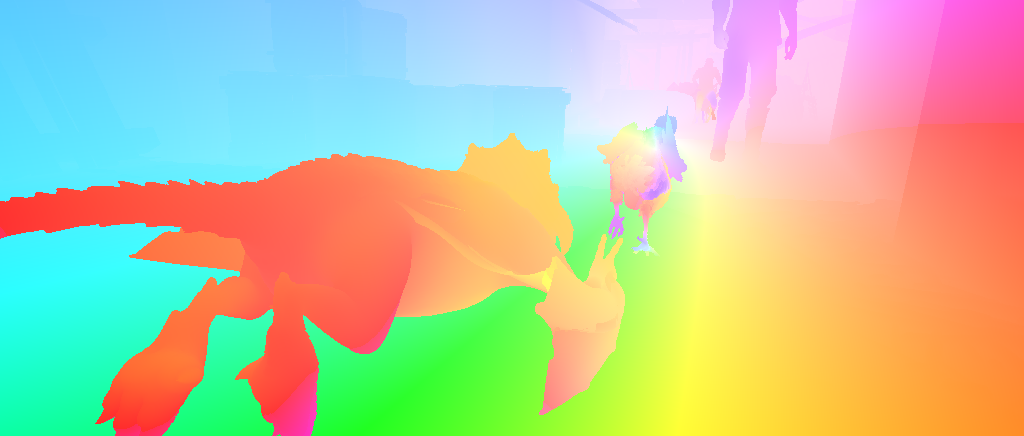}
     \end{subfigure}%
     \hfill
     \begin{subfigure}[b]{0.141\textwidth}
         \centering
         \caption*{Error (RAFT \cite{raft})}
         \includegraphics[width=\textwidth, trim=0 0 250pt 0, clip]{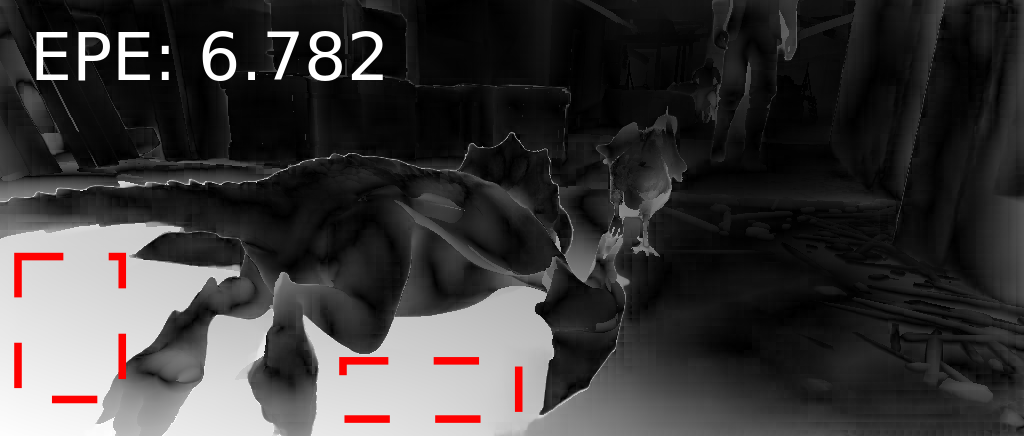}
     \end{subfigure}%
     \hfill
     \begin{subfigure}[b]{0.141\textwidth}
         \centering
         \caption*{Error (Ours)}
         \includegraphics[width=\textwidth, trim=0 0 250pt 0, clip]{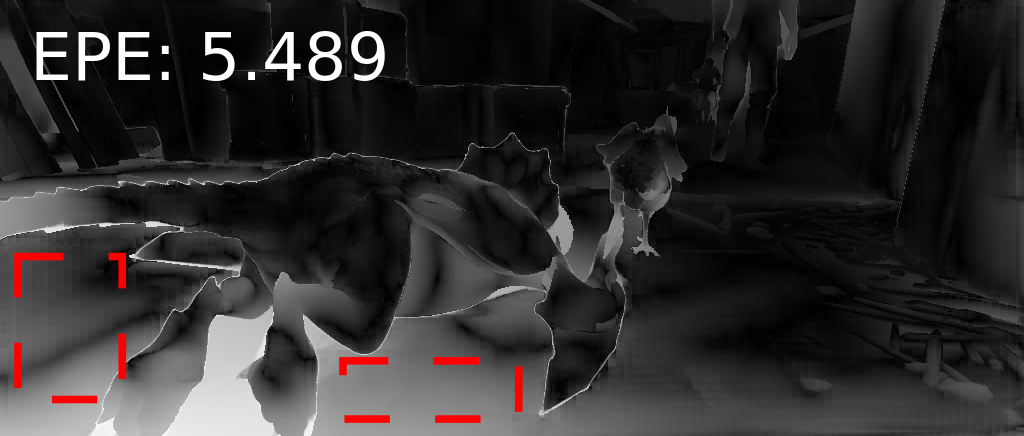}
     \end{subfigure}%
     \vfill
     \begin{subfigure}[b]{0.141\textwidth}
         \centering
         \includegraphics[width=\textwidth, trim=0 0 250pt 0, clip]{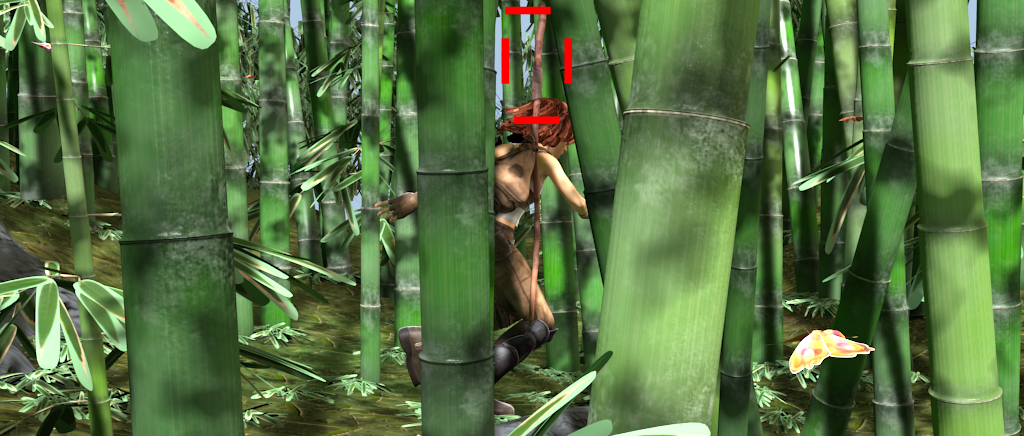}
         \label{fig:input1}
     \end{subfigure}%
     \hfill
     \begin{subfigure}[b]{0.141\textwidth}
         \centering
         \includegraphics[width=\textwidth, trim=0 0 250pt 0, clip]{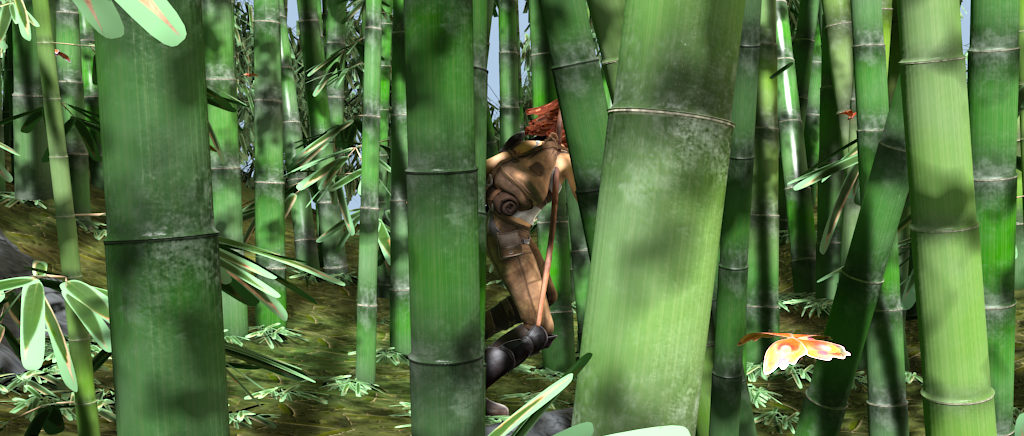}
         \label{fig:input2}
     \end{subfigure}%
     \hfill
     \begin{subfigure}[b]{0.141\textwidth}
         \centering
         \includegraphics[width=\textwidth, trim=0 0 250pt 0, clip]{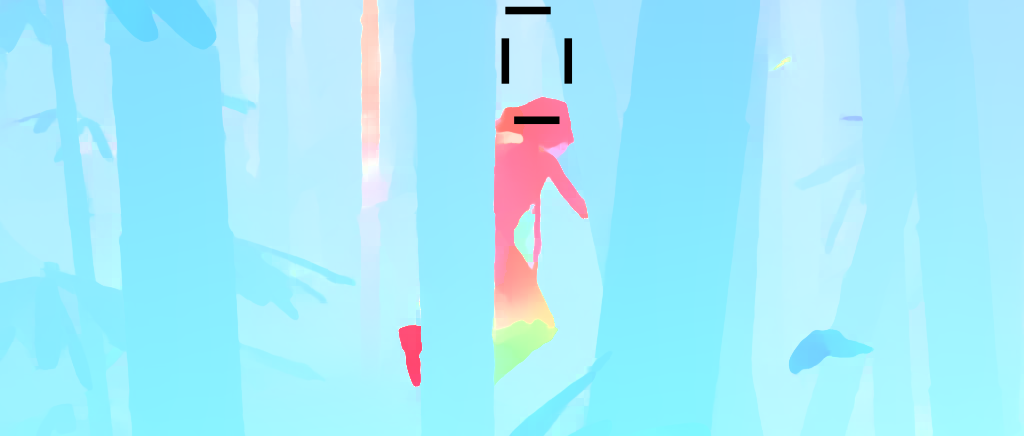}
         \label{fig:raft}
     \end{subfigure}%
     \hfill
     \begin{subfigure}[b]{0.141\textwidth}
         \centering
         \includegraphics[width=\textwidth, trim=0 0 250pt 0, clip]{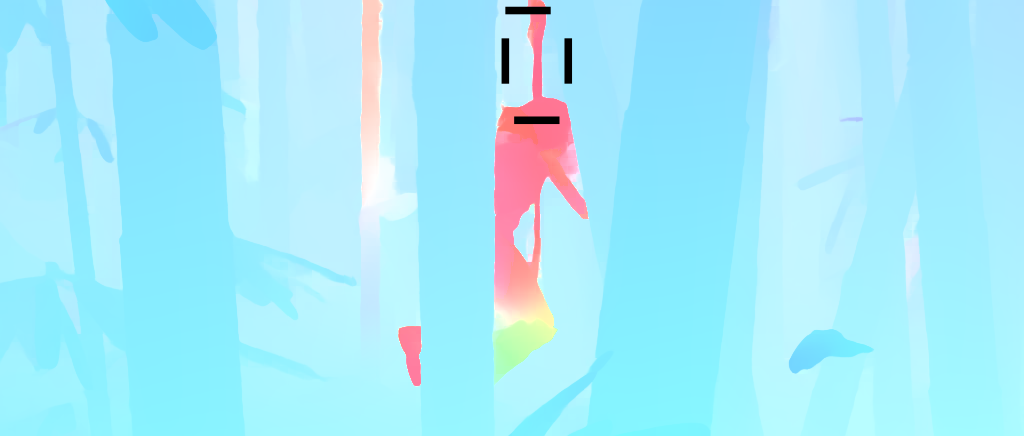}
         \label{fig:ours}
     \end{subfigure}%
     \hfill
     \begin{subfigure}[b]{0.141\textwidth}
         \centering
         \includegraphics[width=\textwidth, trim=0 0 250pt 0, clip]{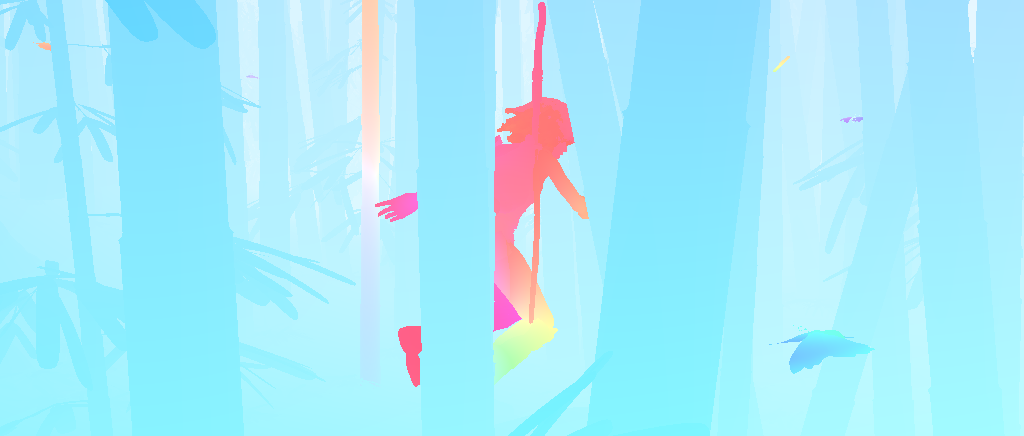}
         \label{fig:ours}
     \end{subfigure}%
     \hfill
     \begin{subfigure}[b]{0.141\textwidth}
         \centering
         \includegraphics[width=\textwidth, trim=0 0 250pt 0, clip]{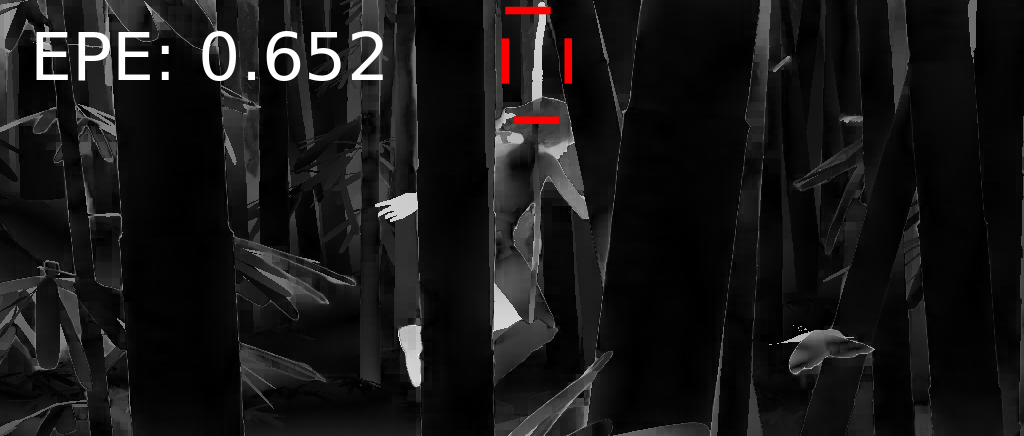}
         \label{fig:ours}
     \end{subfigure}%
     \hfill
     \begin{subfigure}[b]{0.141\textwidth}
         \centering
         \includegraphics[width=\textwidth, trim=0 0 250pt 0, clip]{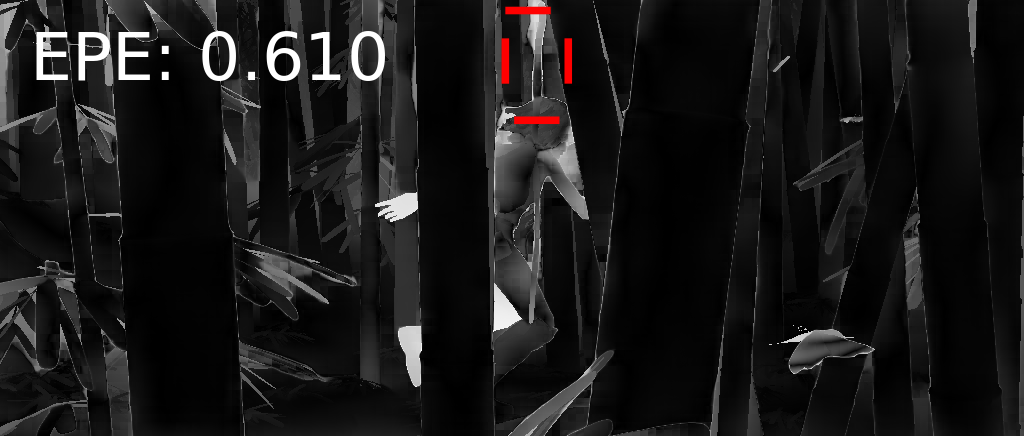}
         \label{fig:ours}
     \end{subfigure}%
     \vspace{-8pt}
     \caption{
     \textbf{Recovering hidden motions.}
     In row 1, the bottom left corner of the ground moves out-of-frame, but reasoning that it
     belongs to the background allows the motion to be recovered from other parts of the image. 
     In row 2, the girl's staff is mostly occluded in the second frame, but strong cues from the 
     visible parts can resolve its motion.
     Our approach can estimate many hidden motions despite the presence of occlusions.
     The flow maps and the error maps have been fetched from the Sintel server \cite{sintel}. 
     Best viewed in colour on a screen. 
     }
    \label{fig:qualitative}
\end{figure*}

The challenge posed by occlusions can be understood by looking at the underlying
assumptions of optical flow algorithms.
Traditional optical flow algorithms apply the brightness constancy constraint \cite{horn},
where pixels related by the flow field are assumed to have the same intensities.
It is clear that occlusions are a direct violation of such a constraint. 
In the deep learning era, correlation (cost) volumes \cite{hosni2012fast} are used to give
a matching cost for each potential displacement of a pixel. However, correlations of appearance
features are unable to give meaningful guidance for learning the motion of occluded regions.
Most existing approaches use smoothness terms in an MRF to interpolate 
occluded motions \cite{fullflow} or use CNNs to directly learn the neighbouring relationships,
hoping to learn to estimate occluded motions based on the neighbouring pixels \cite{raft, pwcnet}.
However, state-of-the-art methods still fail to estimate occluded motions correctly when 
occlusions are more significant and local evidence is insufficient to resolve the ambiguity.

In contrast, humans are able to synthesise information from across the image and apply plausible
motion models to accurately estimate occluded motions. This capability is valuable to emulate,
because we fundamentally care about recovering the real 3D motion of objects in a scene, for which
estimating occluded motion is necessary. Downstream applications, including tracking and activity
detection \cite{tracking}, can also benefit from short-term predictions of the motion of occluded points, particularly
if they reappear later or exhibit some characteristic of interest (\eg, high-velocity out-of-frame motions).

Let us consider how to estimate these hidden motions for the two-frame case. When direct (local)
matching information is absent, the motion information has to be propagated from other pixels.
Using convolutions to propagate this information has the drawback of limited range since
convolution is a local operation.
We propose to aggregate the motion features with a non-local approach. Our design is based on the
assumption that the motions of a single object (in the foreground or background) are often homogeneous. 
One source of information that is overlooked by existing works is self-similarities
in the reference frame. For each pixel, understanding which other pixels are related to it,
or which object it belongs to, is
an important cue for accurate optical flow predictions.
That is, the motion information of non-occluded self-similar points can be propagated to the occluded points.
Inspired by the recent success of transformers \cite{transformer}, we introduce a 
global motion aggregation (GMA) module, where we first compute an attention matrix based on the 
self-similarities of the reference frame, then use that attention matrix to aggregate motion features.
We use these globally aggregated motion features to augment the successful RAFT \cite{raft} framework
and demonstrate new state-of-the-art results in optical flow estimation, such as those examples in Figure~\ref{fig:qualitative}.

The key contributions of our paper are as follows.
We show that long-range connections, implemented using the attention mechanism of transformer networks,
are highly beneficial for optical flow estimation, particularly for resolving the motion of occluded pixels
where local information is insufficient.
We show that self-similarities in the reference frame provide an important cue for selecting the long-range connections
to prioritise.
We demonstrate that our global motion feature aggregation strategy leads to a significant improvement in optical flow accuracy in occluded regions, without damaging the performance in non-occluded regions, and analyse this extensively.
We improve the average end-point error (EPE) by 13.6\% (2.86 $\rightarrow$ 2.47) on Sintel Final and 13.7\% (1.61 $\rightarrow$ 1.39) on Sintel Clean, compared to the strong baseline of RAFT \cite{raft}.
Our approach ranks first on both datasets at the time of submission. 

\begin{figure*}[t!]
    \centering
    \includegraphics[width=\textwidth]{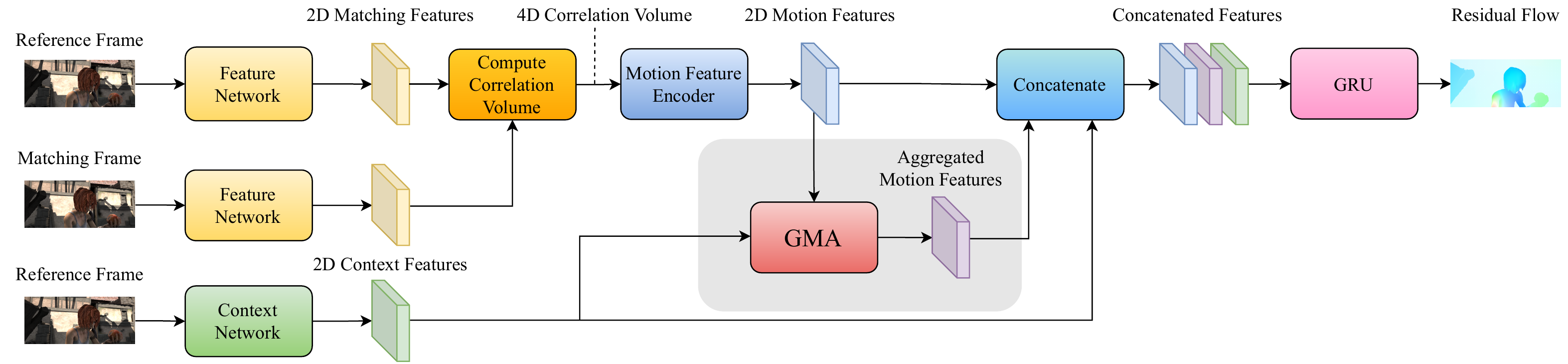}
    \caption{
    \textbf{Proposed architecture.}
    Our network is based on the successful RAFT \cite{raft} architecture.
    The proposed global motion aggregation (GMA) module is contained inside
    the shaded box, a self-contained addition to RAFT with low computational overhead
    that significantly improves performance.
    It takes the visual context features and the motion features as input
    and outputs aggregated motion features that share information across the image.
    These aggregated global motion features are then concatenated with the local motion 
    features and the visual context features to be decoded by the GRU into residual flow.
    This gives the network the flexibility to choose between or combine the local and global
    motion features, depending on the needs of the specific pixel location. For example,
    a location with poor local image evidence, caused by occlusion for instance, could preference the
    global motion features.
    }
    \label{fig:overall}
\end{figure*}

\section{Related Works}
\label{Sec:related}

\paragraph{Occlusions in optical flow.}

Occlusion poses a key challenge in optical flow estimation due to its violation of the brightness
constancy constraint \cite{horn}. Most traditional optical flow algorithms treat occlusions as outliers
and so develop and optimise robust objective functions. In continuous optimisation for optical flow, 
Brox \etal \cite{broxl1} used the $L^1$ norm due to its robustness to outliers caused by occlusions
or large brightness variations. Zach \etal \cite{tvl1} added total variation regularisation
and proposed an efficient numerical scheme to optimise the energy functional. 
This formulation was later improved by Wedel \etal \cite{wedel2009improved}. Later work introduced
additional robust optimisation terms, including the Charbonnier potential \cite{LKmeetsHS} and the Lorentzian
potential \cite{lorentzian}.

More recently, discrete optimisation approaches, especially Markov Random Fields (MRFs) \cite{boykov},
have been used to estimate optical flow. These algorithms \cite{discreteflow, fullflow, dcflow} first estimate
the forward and backward flows separately using a robust, truncated data term. They then conduct a 
forward--backward consistency check to determine the occluded regions. Lastly, as a post-processing step,
they use interpolation methods \cite{epicflow} to fill in the optical flow of the occluded regions.

Other work incorporates occlusion estimation as a joint objective together with optical flow
estimation. Alvarez \etal \cite{alvarez} use
forward--backward consistency as an optimisation objective, thus estimating time-symmetric
optical flow. In addition to forward--backward consistency, MirrorFlow \cite{mirrorflow} 
incorporates occlusion--disocclusion symmetry in the energy function and achieves
performance improvements. Since occlusions are caused by 3D motions, other works
\cite{sun2010layered, sun2014local} explicitly model 
local depth relationships into layers and reason about occlusions. 

Contrary to the above approaches, we do not overload the loss function with explicit occlusion reasoning.
Instead, we adopt a learning approach, similar to other supervised deep optical flow learning
approaches \cite{pwcnet,hd3,liteflownet,irr,vcn,scopeflow,maskflownet,raft,scv}. Rather than estimating
an occlusion map explicitly, our goal is to improve the optical flow accuracy at occluded regions. 
We take an implicit approach to globally aggregate motion features, which provides extra information
to correctly predict flow at occluded regions. Our approach can be thought of as a non-local interpolation
approach \cite{sun2010secrets}, in contrast to local interpolation approaches \cite{epicflow}. 
In the deep learning literature, the occlusion problem
has been addressed in an unsupervised learning setting \cite{occaware,multiframe,selflow}, 
however, existing supervised learning approaches all rely on convolutions
to interpolate in occluded regions, which are prone to failure for
more significant occlusions. 

\paragraph{Self-attention and transformers.}
Our design principle is inspired by the recent successes of the transformer literature \cite{transformer}. 
The transformer architecture was first successful in natural language processing (NLP), 
due to its ability to model long-range dependencies and its scalability for GPU parallel processing. 
Among various modules in the transformer achitecture, self-attention is the key design feature that make transformers
work. Recently, researchers have introduced the transformer and related attention ideas to the vision 
community, mostly in high-level tasks such as image classification \cite{standalone, vit} 
and semantic segmentation \cite{dual, nonlocal, crisscross}. 
To the best of our knowledge, we are the first to use the idea of attention
to solve the optical flow problem. Different from many existing works in the transformer
literature, we do not use self-attention in our work. 
Self-attention refers to the query, key and value vectors coming from the same features. In our case, 
query and key vectors come from the context features modelling the appearance of the image while 
value vectors come from the motion features, which is an encoding of the correlation volume.  

\begin{figure*}[t!]
    \centering
    \includegraphics[width=0.7\textwidth]{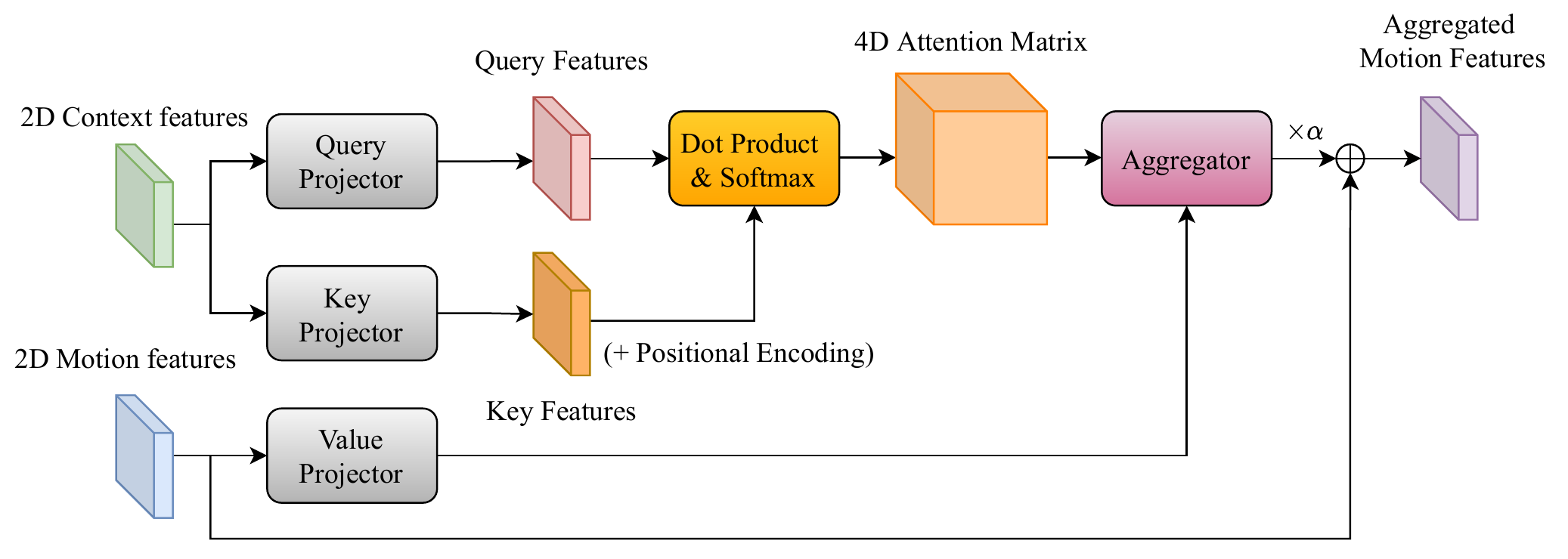}
    \caption{
    \textbf{Details of the GMA module.}
    To model the self-similarity of the first frame, we 
    project the context feature map to a query feature map and a key feature map. We then
    take the dot product of the two feature maps and a softmax to obtain an attention matrix, which 
    encodes self-similarity in appearance feature space. Similar to transformer networks \cite{transformer}, we also take the 
    dot product between the query feature map and a set of positional embedding vectors 
    which augments the attention matrix with positional information.       
    Separately, the motion feature map encoded from the correlation volume is projected using the learned value
    projector. Its weighted sum, using the obtained attention matrix, produces the aggregated global motion features.}
    \label{fig:details}
\end{figure*}
\section{Method}
\label{Sec:method}

\subsection{Background}
We base our network design on the successful RAFT architecture \cite{raft}. Our overall network diagram
is shown in Figure~\ref{fig:overall}. For completeness, we briefly describe the main contributions
of RAFT from which our model benefits. The first
contribution is the introduction of an all-pairs correlation volume, which explicitly models matching 
correlations for all possible displacements. The benefit of using all-pairs correlations is its ability
to handle large motions. The second major contribution is the use of a gated recurrent unit (GRU) decoder for iterative residual
refinement \cite{irr}. The constructed 4D correlation volumes are encoded to 2D motion features, which
are iteratively decoded to predict the residual flow. The final flow prediction is a sum of the sequence
of residual flows. The benefit of using a GRU to perform iterative refinement lies in the reduction of the search
space. In RAFT, convolutions are used in the GRU decoder, which learn to model spatial smoothness. Due to
the local nature of convolutions, they can learn to handle small occlusions but tend to fail when these become
more significant and local evidence is insufficient to resolve the motion. 

\subsection{Overview}
In his first paper from 1976, Geoffrey Hinton wrote that ``local ambiguities have to be 
resolved by finding the best global interpretation'' \cite{hinton1976using}. This idea
still holds true in the modern deep learning era. To resolve ambiguities caused by
occlusions, our core idea is to allow the network to reason at a higher level, that is,
to globally aggregate the motion features of similar pixels, having implicitly reasoned about
which pixels are similar in appearance feature space.
We hypothesise that the network will be able to find points with similar motions by looking for points with similar appearance in the reference frame.
This is motivated by the observation that the motions of points on a single object are often homogeneous.
For example, the motion vectors of a person running to the right have a bias towards the right, which holds even if we do not see where a large part of the person ends up in the matching frame due to occlusion.
We can use this statistical bias to propagate motion information from non-occluded pixels,
with high (implicit) confidence, to occluded pixels, with low confidence. Here, confidence
can be interpreted as whether there exists a distinct matching, \ie, a high correlation value at the correct displacement. 

With these ideas, we take inspiration from transformer networks \cite{transformer}, which are known
for their ability to model long-range dependencies. Different from the self-attention mechanism in transformers,  
where query, key and value come from the same feature vectors, we use a generalized variant of attention.
Our query and key features are projections of the context feature map, which are used to model the
appearance self-similarities in frame 1. The value features are projections of the motion features, which
themselves are an encoding of the 4D correlation volume. The attention matrix computed from the query and key
features is used to aggregate the value features which are hidden representations of motions. We name this a Global Motion Aggregation (GMA) module. The aggregated motion features are concatenated with
the local motion features as well as the context features, which is to be decoded by the GRU. 
A detailed diagram of GMA is shown in Figure~\ref{fig:details}.

\begin{figure*}[t!]
     \centering
     \begin{subfigure}[b]{0.19\textwidth}
         \centering
         \includegraphics[width=\textwidth]{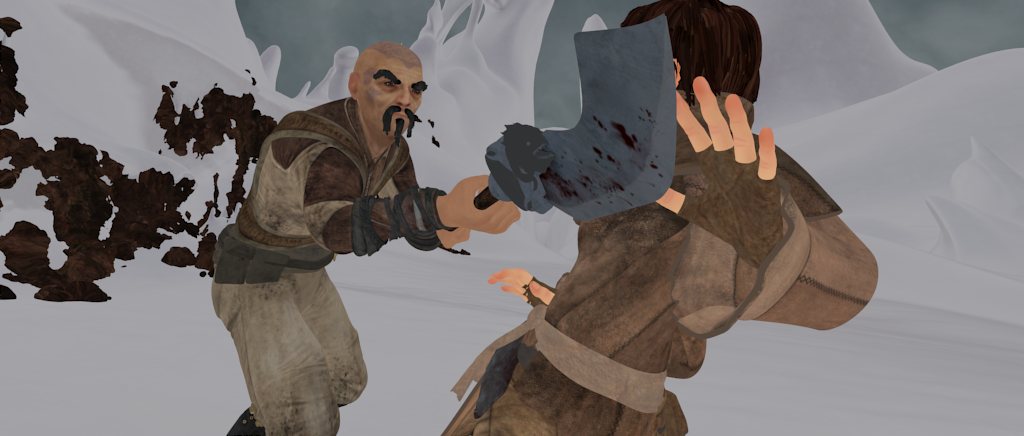}
         \caption{Reference Frame}
         \label{fig:input1}
     \end{subfigure}
     \hfill
     \begin{subfigure}[b]{0.19\textwidth}
         \centering
         \includegraphics[width=\textwidth]{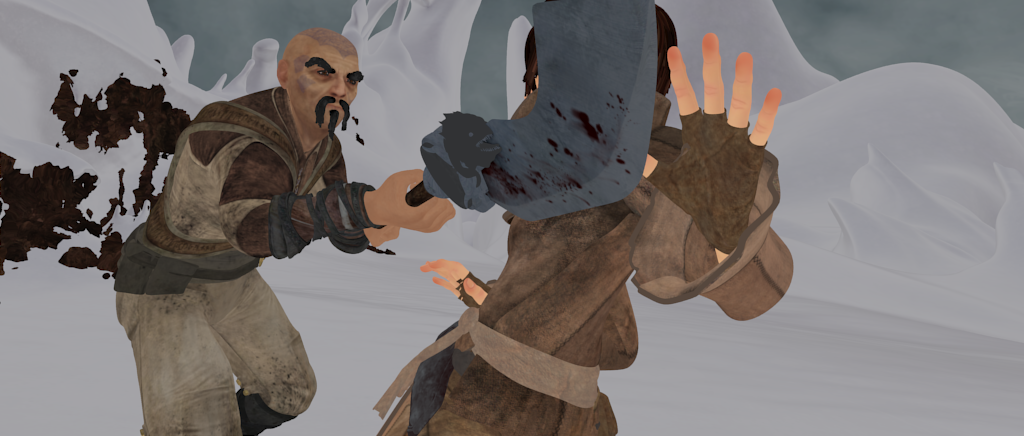}
         \caption{Matching Frame}
         \label{fig:input2}
     \end{subfigure}
    \hfill 
     \begin{subfigure}[b]{0.19\textwidth}
         \centering
         \includegraphics[width=\textwidth]{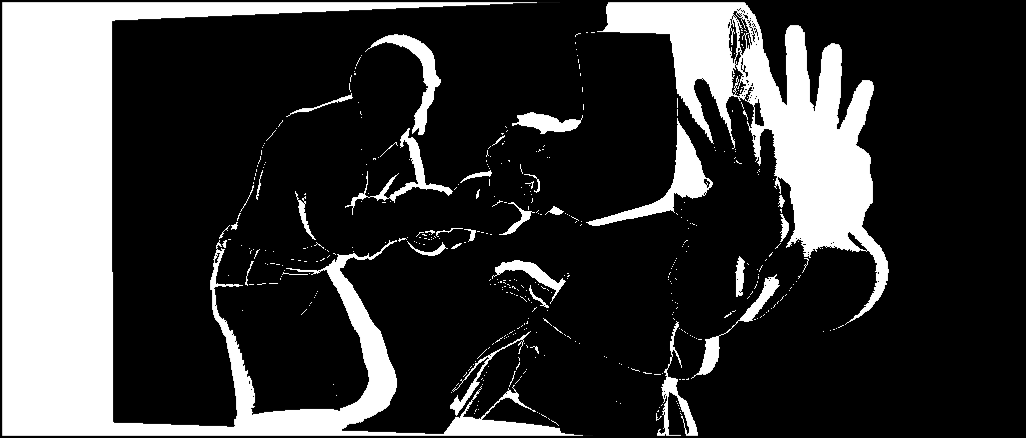}
         \caption{Occ}
         \label{fig:occ1}
     \end{subfigure}
    \hfill 
     \begin{subfigure}[b]{0.19\textwidth}
         \centering
         \includegraphics[width=\textwidth]{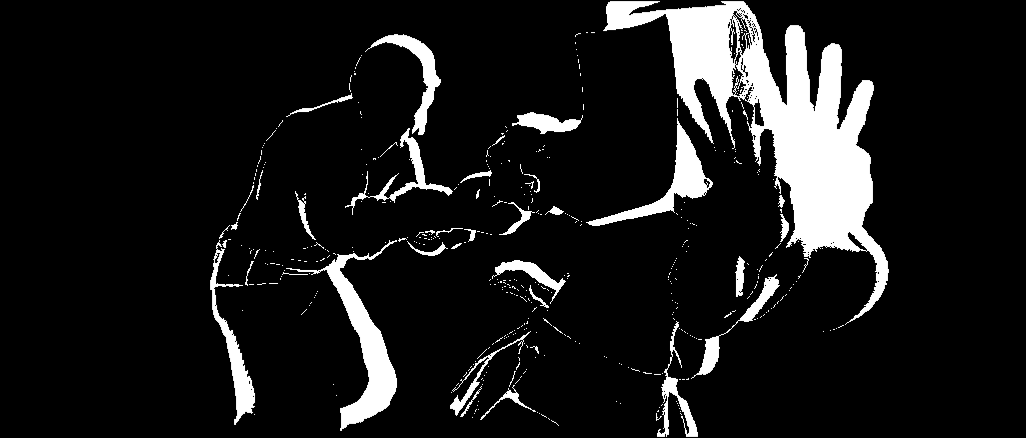}
         \caption{Occ-in}
         \label{fig:occ2}
     \end{subfigure}
    \hfill 
     \begin{subfigure}[b]{0.19\textwidth}
         \centering
         \includegraphics[width=\textwidth]{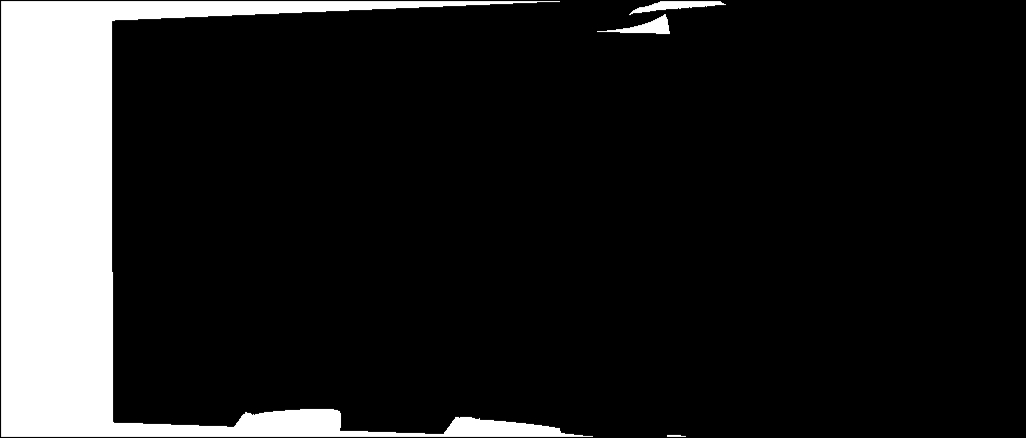}
         \caption{Occ-out}
         \label{fig:occ3}
     \end{subfigure}
     \caption{\textbf{Examples of the Sintel Albedo dataset and occlusion maps.} The 
     Albedo dataset is rendered without the illumination effects. The occlusion map in
     this example contains mostly foreground objects occluding the background scene
     as well as the background on the left moving out of the field-of-view. Figure~\ref{fig:occ1}
     is the occlusion map (Occ) for this example. 
     Figure~\ref{fig:occ2} and Figure~\ref{fig:occ3}
     are the in-frame (Occ-in) and out-of-frame (Occ-out) occlusion maps respectively.}
    \label{fig:albedo}
\end{figure*}

\subsection{Mathematical Formulation}
Let ${\rvx \in \R^{N \times D_\text{c}}}$ denote the context (appearance) features and 
${\rvy \in \R^{N \times D_\text{m}}}$ 
denote the motion features, where ${N = HW}$ and $H$ and $W$ are the height
and width of the feature map, $D$ refers to the channel dimension of the feature map. 
The $i^{\text{th}}$ feature vector is denoted ${\rvx_i \in \R^{D_\text{c}}}$. 
Our GMA module computes the feature vector update as an attention-weighted sum
of the projected motion features. The aggregated motion features are given by
\begin{equation}
    \hat{\rvy}_i = \rvy_i + \alpha \sum_{j=1}^N f(\theta(\rvx_i), \phi(\rvx_j))\sigma(\rvy_j),
    \label{Eqn:c_only}
\end{equation}
where $\alpha$ is a learned scalar parameter initialised to zero,
$\theta$, $\phi$ and $\sigma$ are the projection functions for the query, key, and value vectors,
and $f$ is a similarity attention function given by
\begin{equation}
    f(\rva_i, \rvb_j) = \frac{\exp \left( \rva_i\transpose \rvb_j / \sqrt{D} \right)}{\sum_{j=1}^N \exp \left( \rva_i\transpose \rvb_j / \sqrt{D} \right)}.
\end{equation}
The projection functions for the query, key and value vectors are given by
\begin{align}
   \theta (\rvx_i) = \rmW_{\text{qry}} \rvx_i, \\
   \phi (\rvx_i) = \rmW_{\text{key}} \rvx_i, \\
   \sigma (\rvy_i) = \rmW_{\text{val}} \rvy_i,
\end{align}
where $\rmW_{\text{qry}}, \rmW_{\text{key}} \in \R^{D_\text{in} \times D_\text{c}}$ and 
$\rmW_{\text{val}} \in \R^{D_\text{m} \times D_\text{m}}$. The learnable parameters in our
GMA module include $\rmW_{\text{qry}}, \rmW_{\text{key}}, \rmW_{\text{val}}$ and $\alpha$. 

The final output is $[\rvy \,|\, \hat{\rvy} \,|\, \rvx]$, a concatenation of the three feature maps.
The GRU decodes this to obtain the residual flow.
Concatenation allows the network to intelligently select from or combine the motion vectors,
modulated by the global context feature, without prescribing exactly how it is to do this.
It is plausible that the network learns to encode some notion of uncertainty, and decodes
the aggregated motion vector only when the model cannot be certain of the flow
from the local evidence.

We also explore the use of a 2D relative positional embedding \cite{bello2019attention}, allowing the attention map
to depend on both the feature self-similarity and the relative position from the query point.
For this, we compute the aggregated motion vector as
\begin{equation}
    \hat{\rvy}_i = \rvy_i + \alpha \sum_{j=1}^N f(\theta(\rvx_i), \phi(\rvx_j) + \rvp_{j-i}) \sigma(\rvy_j),
    \label{Eqn:p+c}
\end{equation}
where $\rvp_{j-i}$ denotes the relative positional embedding vector indexed by the pixel offset $j-i$.
Separate embedding vectors are learned for the vertical and horizontal offsets and are summed to obtain $\rvp_{j-i}$.
If it is useful to suppress pixels that are very close or very far from the query point when aggregating the motion vectors,
then this positional embedding has the capacity to learn this behaviour.

We also investigated computing the attention map from only the query vectors and
positional embedding vectors, without any notion of self-similarity. That is,
\begin{equation}
    \hat{\rvy}_i = \rvy_i + \alpha \sum_{j=1}^N f(\theta(\rvx_i), \rvp_{j-i}) \sigma(\rvy_j).
    \label{Eqn:p_only}
\end{equation}
This can be regarded as learning long-range aggregation without reasoning about the image content.
It is plausible that positional biases in the dataset could be exploited by such a scheme.
In Table~\ref{Tab:Results}, the results for (\ref{Eqn:p+c}) and (\ref{Eqn:p_only}) are denoted as
Ours~(+p) and Ours~(p only).

\section{Experiments}
\label{Sec:experiments}
\subsection{Experimental Setup}
\label{Sec:details}

We follow the standard optical flow training procedure \cite{flownet2,pwcnet,raft}
of first pre-training our model on FlyingChairs
\cite{flownet} for 120k iterations with a batch size of 8 and then on FlyingThings
\cite{things} for another 120k iterations with a batch size of 6. We then fine-tune on
a combination of FlyingThings, Sintel \cite{sintel}, KITTI 2015 \cite{kitti} and HD1K \cite{hd1k} 
for 120k iterations for Sintel evaluation and 50k on KITTI 2015 \cite{kitti} for KITTI evaluation. 
A batch size of 6 is set for fine-tuning. We train our model on two 2080Ti GPUs with the 
\texttt{PyTorch} library \cite{pytorch} using the mixed precision strategy.
We adopt the same hyperparameters as RAFT \cite{raft} for the base network. 
We adopt the one-cycle learning rate policy \cite{onecycle} with the highest learning rate
set to $2.5 \times 10^{-4}$ for FlyingChairs then $1.25 \times 10^{-4}$ for the rest. 
For GMA, we choose channel dimensions $D_{\text{in}} = D_{\text{c}} = D_{\text{m}} = 128.$

The main evaluation metric we use is average end-point error (AEPE), which refers to the mean
pixelwise flow error. KITTI also uses the Fl-All (\%) metric which refers to the percentage
of optical flow vectors whose end-point error is larger than 3 pixels or over $5\%$ of
ground truth. 

The Sintel dataset has been created with different rendering passes that have different levels of complexity.
For training and test evaluation on the Sintel server, we used the Clean and Final passes.
The Clean pass is rendered with illumination including smooth shading and specular reflections. The Final pass 
is created with full rendering, which includes motion blur, camera depth-of-field blur, and atmospheric effects. 

In the Sintel training set, they also provided the Albedo pass, which is rendered without illumination effects
and has roughly piecewise-constant colours. An example is shown in Figure~\ref{fig:albedo}. 
We do not use this set for training, but reserve it as an evaluation dataset.
The motivation for doing so is that the Albedo set adheres to brightness constancy everywhere 
apart from occluded regions. 
By evaluating and analysing on the occluded regions and non-occluded regions separately, 
we can clearly see how well our method performs when addressing the occlusion problem. 

\subsection{Occlusion Analysis}
\setlength\tabcolsep{4pt}
\begin{table}[t!]
\centering
\newcolumntype{C}{>{\centering\arraybackslash}X}
\begin{tabularx}{\columnwidth}{l l C C C}
\toprule
Sintel & & RAFT & Ours & Rel. Impr.\\
Dataset & Type & (AEPE) & (AEPE) & (\%)\\
\midrule
\multirow{5}{1.2cm}{Clean (train)} 
                            & Noc & 0.32 & \textbf{0.29} & 9.3 \\
                            & Occ & 5.36 & \textbf{4.25} & 20.7 \\
                            & Occ-in & 4.45 & \textbf{3.81} & 14.4 \\
                            & Occ-out & 7.01 & \textbf{5.03} & \textbf{28.2} \\
                            & All & 0.74 & \textbf{0.62} & 16.2 \\
\midrule
\multirow{5}{1.2cm}{Final (train)} 
                            & Noc & 0.66 & \textbf{0.59} & 10.6 \\
                            & Occ & 7.09 & \textbf{6.22} & 12.2 \\
                            & Occ-in & 6.21 & \textbf{5.30} & \textbf{14.6} \\
                            & Occ-out & 8.71 & \textbf{7.90} & 9.3 \\
                            & All & 1.19 & \textbf{1.06} & 10.9 \\
\midrule
\multirow{5}{1.2cm}{Albedo (test)}
                            & Noc & 0.34 & \textbf{0.32} & 5.9 \\
                            & Occ & 6.35 & \textbf{5.58} & 12.1 \\ 
                            & Occ-in & 5.83 & \textbf{5.23} & 10.3 \\ 
                            & Occ-out & 7.29 & \textbf{6.20} & \textbf{15.0} \\ 
                            & All & 0.84 & \textbf{0.76} & 9.5 \\
\bottomrule
\end{tabularx}
\vspace{0pt} 
\caption{
\textbf{Optical flow error for different Sintel datasets}, partitioned into occluded 
(`Occ') and non-occluded (`Noc') regions. In-frame and out-of-frame occlusions
are further split and denoted as `Occ-in' and `Occ-out'. The best results and the largest
relative improvement in each dataset are styled in bold.}
\label{Tab:Occ}
\end{table}

To verify the effectiveness of our proposed GMA module at estimating the motion of occluded points,
we make use of the occlusion maps provided in the Sintel training set, which partition the pixels into non-occluded (Noc) and occluded (Occ) pixels. We further divide the occluded pixels into in-frame
(`Occ-in') and out-of-frame (`Occ-out') occlusions, depending on whether the 
ground-truth flow vector points inside or outside the image frame.
An example is shown in Figure~\ref{fig:albedo}.

We evaluated on all three rendering passes of Sintel, where the results for Clean and Final are training set errors and those for Albedo are test set errors. 
We evaluated the AEPE for different regions, results of which are shown in Table~\ref{Tab:Occ}.
We observe that the relative improvement of our method compared to RAFT is predominantly 
attributable to better predictions of the flow for occluded points. 
This is reinforced by the results on the Albedo dataset where the brightness constancy assumption holds exactly for non-occluded points, removing confounding factors.
Finally, out-of-frame occlusions are more challenging than in-frame occlusions for both models, but we still observe a significant improvement for these pixels.
We hypothesise that the improvement in non-occluded regions is due to GMA's ability to resolve ambiguities
caused by other brightness variations, for example specular reflections, blurs, and other sources.
This result strongly supports our claim that global aggregation can help resolve ambiguities caused by occlusion. 

\subsection{Comparison with Prior Works}
\label{Sec:quantitative}

Having shown that our approach can improve optical flow estimates for occluded regions, we compare
against prior works on the overall performance. 
We evaluate our approach on the Sintel dataset \cite{sintel} and the KITTI 2015 optical flow dataset
\cite{kitti}. 
At the time of submission, we have obtained the best results on both the Sintel Final and Clean benchmarks 
among all submitted results published and unpublished. 
Compared with our baseline approach RAFT \cite{raft}, we have improved the AEPE from $2.86$
to $2.47$ ($13.6\%$ improvement) on Sintel Final and $1.61$ to $1.39$ ($13.7\%$ improvement) on Sintel Clean. 
This significant improvement over RAFT validates our claim that our approach can improve 
flow prediction for occluded regions without damaging the performance of non-occluded regions. 
The Sintel server also reports the metric `EPE unmatched', which measures the endpoint error over regions
that are visible only in one frame, predominantly caused by occlusion. Our approach also ranks first
under this metric in both Clean and Final, with a margin of 0.9 EPE on Clean (2.2 \wrt RAFT) and 1.3 EPE on Final (1.7 \wrt RAFT). 
Overall, our model achieves a new state-of-the-art result in optical flow estimation, which demonstrates
the usefulness of addressing the occlusion problem in optical flow.

On the KITTI 2015 test set, our results are on par with RAFT. `Ours~(p only)', which uses positional attention
only, outperforms RAFT, while `Ours', which uses content self-similarity attention, slightly underperforms.
It is likely that the lack of improvement on this dataset is due to having insufficient
training data (only 200 pairs of images) for the network to learn high-level appearance feature similarities.

\setlength\tabcolsep{4pt}
\begin{table*}[t!]
\centering
\newcolumntype{C}{>{\centering\arraybackslash}X}
\begin{tabularx}{\textwidth}{@{}l l C C C C C C c@{}}
\toprule
\multirow{2}[3]{1.5cm}{Training Data} & & \multicolumn{2}{c}{Sintel (train)} & \multicolumn{2}{c}{KITTI-15 (train)} & \multicolumn{2}{c}{Sintel (test)} & \multicolumn{1}{c}{KITTI-15 (test)}\\
\cmidrule(lr){3-4}
\cmidrule(lr){5-6}
\cmidrule(lr){7-8}
\cmidrule(lr){9-9}
& Method & Clean & Final & AEPE & Fl-all (\%) & Clean & Final & Fl-all (\%)\\
\midrule
\multirow{9}{*}{C + T} 
                       & VCN\cite{vcn}            & 2.21  & 3.68  & 8.36 & 25.1 & - & -     & - \\ 
                       & MaskFlowNet\cite{maskflownet} & 2.25 & 3.61 & - & 23.1 & - & - & - \\ 
                       & FlowNet2\cite{flownet2}       & 2.02  & 3.54 & 10.08 & 30.0 & 3.96  & 6.02 & - \\
                       & RAFT\cite{raft}        & 1.43 & \textbf{2.71} & 5.04 & 17.4 & - & - & - \\ 
                       \cmidrule[\lightrulewidth](r{0.3em}){2-9}
                       & Ours (p only) & 1.48 & 2.88 & 5.01 & 16.9 & - & - & - \\
                       & Ours (+p) & 1.33 & 2.87 & 4.83 & \textbf{16.6} & - & - & - \\
                       & Ours & \textbf{1.30} & 2.74 & \textbf{4.69} & 17.1 & - & - & - \\
                       \midrule
\multirow{13}{1.5cm}{C + T + S/K (+ H)} 
                     & FlowNet2 \cite{flownet2}  & (1.45) & (2.01) & (2.30) & (6.8) & 4.16  & 5.74 & 11.48  \\
                     & PWC-Net+\cite{pwcnet+}   & (1.71)     & (2.34)  & (1.50) & (5.3)  & 3.45  & 4.60 & 7.72 \\
                     & VCN \cite{vcn}            & (1.66)     & (2.24) & (1.16) & (4.1) & 2.81  & 4.40 & 6.30 \\
                     & MaskFlowNet\cite{maskflownet} & - & - & - & - & 2.52 & 4.17 & 6.10 \\
                     & RAFT\cite{raft}  & (0.76) & (1.22) & (0.63) & (1.5) & 1.61$^\star$ & 2.86$^\star$  & 5.10 \\
                     \cmidrule[\lightrulewidth](r{0.3em}){2-9}
                     & Ours (p only)  & (0.64) & (1.08) & (\textbf{0.56}) & (\textbf{1.2}) & 1.48$^\star$ & 2.56$^\star$ & \textbf{4.93} \\
                     & Ours (+p)  & (0.65) & (1.11) & (0.58) & (1.3) & 1.54$^\star$ & 2.63$^\star$ & 5.08 \\
                     & Ours  & (\textbf{0.62}) & (\textbf{1.06}) & (0.57) & (\textbf{1.2}) & \textbf{1.39}$^\star$ & \textbf{2.47}$^\star$ & 5.15 \\
                     \bottomrule
\end{tabularx}
\vspace{0pt} 
\caption{\textbf{Quantitative results on Sintel and KITTI 2015 datasets.} We report the average end-point error (AEPE) where not otherwise stated, as well as the Fl-all measure for the KITTI dataset, which is the percentage of optical flow outliers with an error larger than 3 pixels. ``C + T'' refers to results that are pre-trained on the Chairs and Things datasets. ``S/K (+ H)'' refers to methods that are fine-tuned on the Sintel and KITTI datasets, with some also fine-tuned on the HD1K dataset. Parentheses denote training set results and bold font denotes the best result. ``Ours~(p only)'' denotes the position-only attention model defined in (\ref{Eqn:p_only}). ``Ours~(+p)'' denotes the joint position and content-wise attention model defined in (\ref{Eqn:p+c}). ``Ours'' denotes our main content-only self-similarity attention model defined in (\ref{Eqn:c_only}). $^\star$Results evaluated with the ``warm-start'' strategy detailed in the RAFT paper \cite{raft}.
}
\label{Tab:Results}
\end{table*}

\subsection{Qualitative Results}
\label{Sec:qualitative}
\begin{figure*}[ht!]
     \centering
     \begin{subfigure}[b]{0.195\textwidth}
         \centering
         \caption*{Reference Frame}
         \includegraphics[width=\textwidth]{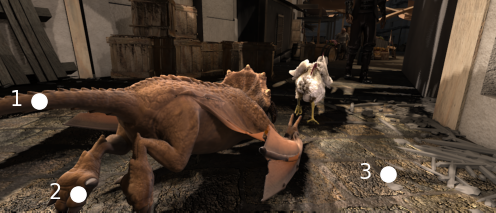}
     \end{subfigure}%
     \hfill
     \begin{subfigure}[b]{0.195\textwidth}
         \centering
         \caption*{Attention map 1}
         \includegraphics[width=\textwidth]{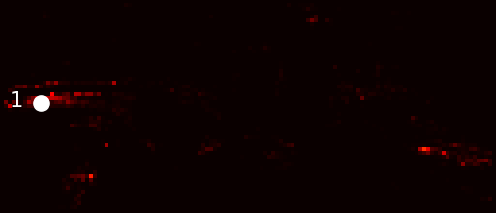}
     \end{subfigure}%
     \hfill
     \begin{subfigure}[b]{0.195\textwidth}
         \centering
         \caption*{Attention map 2}
         \includegraphics[width=\textwidth]{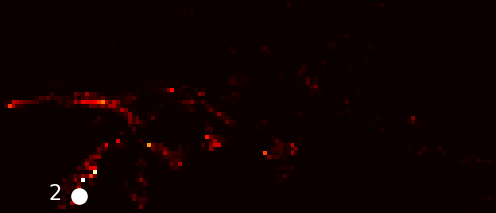}
     \end{subfigure}%
     \hfill
     \begin{subfigure}[b]{0.195\textwidth}
         \centering
         \caption*{Attention map 3}
         \includegraphics[width=\textwidth]{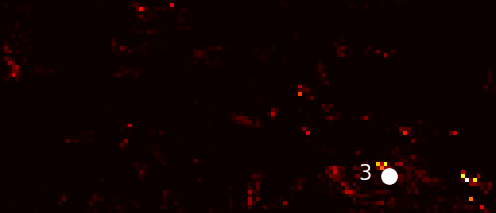}
     \end{subfigure}%
     \hfill
     \begin{subfigure}[b]{0.195\textwidth}
         \centering
         \caption*{Flow}
         \includegraphics[width=\textwidth]{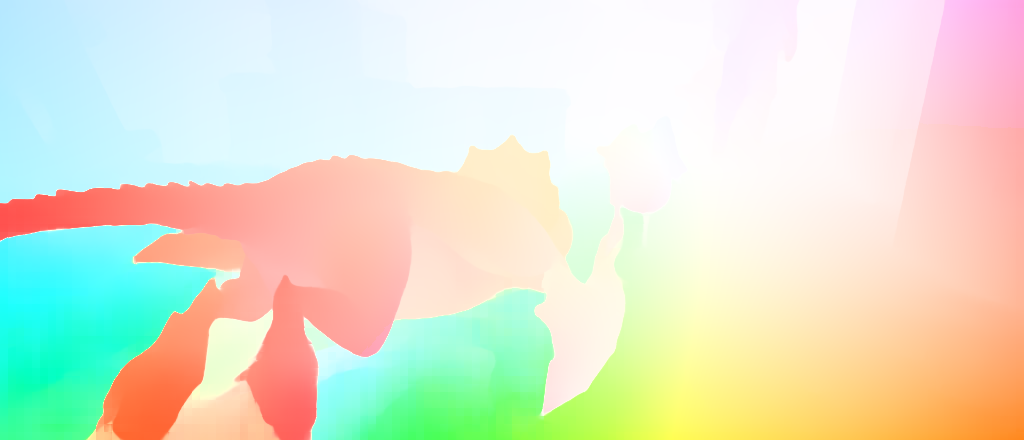}
     \end{subfigure}%
     
     \begin{subfigure}[b]{0.195\textwidth}
         \centering
         \includegraphics[width=\textwidth]{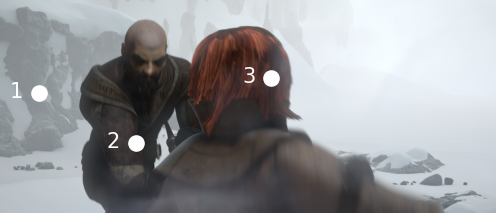}
     \end{subfigure}%
     \hfill
     \begin{subfigure}[b]{0.195\textwidth}
         \centering
         \includegraphics[width=\textwidth]{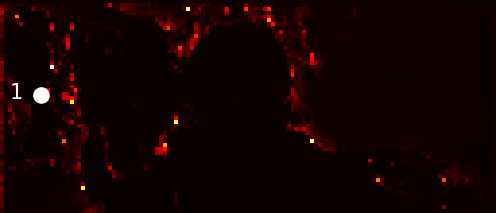}
     \end{subfigure}%
     \hfill
     \begin{subfigure}[b]{0.195\textwidth}
         \centering
         \includegraphics[width=\textwidth]{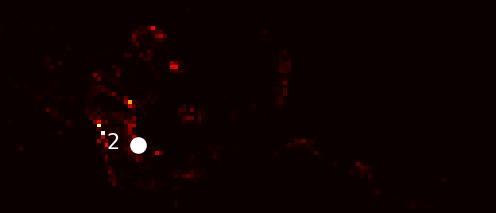}
     \end{subfigure}%
     \hfill
     \begin{subfigure}[b]{0.195\textwidth}
         \centering
         \includegraphics[width=\textwidth]{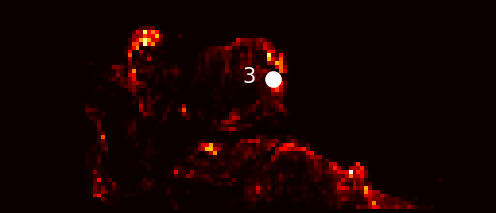}
     \end{subfigure}%
     \hfill
     \begin{subfigure}[b]{0.195\textwidth}
         \centering
         \includegraphics[width=\textwidth]{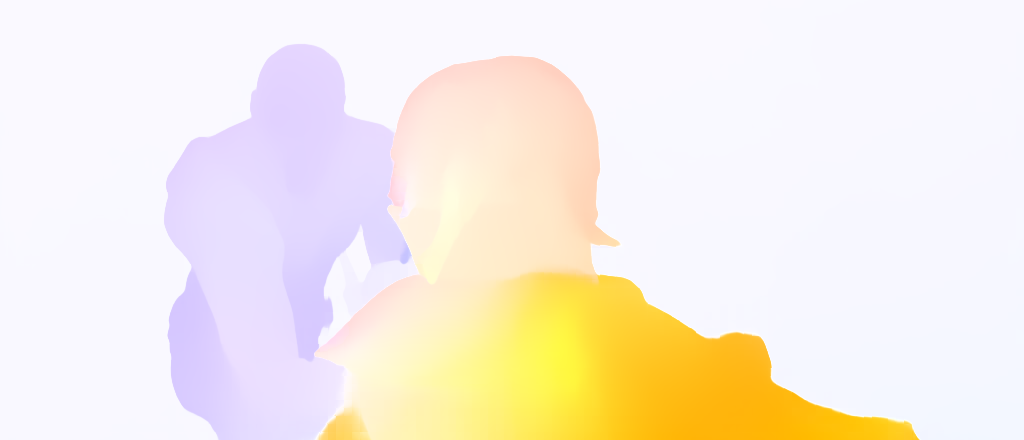}
     \end{subfigure}%
     \caption{\textbf{Attention map visualisations.} For each row, we show
     the first frame and three query points. Then we show the three attention
     maps corresponding to these query points (brighter colours mean higher attention
     weights) as well as the predicted optical flow. 
     }
     \label{fig:attention_maps}
\end{figure*}
Qualitative results are shown in Figure~\ref{fig:qualitative} for two examples in the Sintel Clean dataset. The optical flow error in regions of the image that move out-of-frame or behind another object is significantly reduced compared to RAFT. These scenes are highly challenging with lots of motion and occlusion. For example, it is not unreasonable that RAFT is unable to keep track of the wooden staff that becomes partially occluded in the second image, given that it is well-camouflaged in a forest, fast-moving, and very thin. However, our model is able to very accurately predict the staff's motion, despite these challenges.

We also present visualisations of the learned attention maps for two examples in Figure~\ref{fig:attention_maps}. To train effectively, the network should learn to attend to pixels that share similar motion vectors. For foreground points, we expect this to be  most easily achieved by attending to points on the same object, while for background points it may be sufficient to attend to any other background point. These examples justify this expectation and provide support for the argument that appearance (and higher-order) self-similarity is being learned by the network, and that this is helpful for estimating the flow of the occluded points.

\subsection{Ablation Results}
\label{Sec:ablation}
\setlength\tabcolsep{4pt}
\begin{table}[t!]
\centering
\newcolumntype{C}{>{\centering\arraybackslash}X}
\begin{tabularx}{\columnwidth}{l C C C C C}
\toprule
 & Chairs & \multicolumn{2}{c}{Things} & \multicolumn{2}{c}{Sintel}\\
 & & Clean & Final & Clean & Final \\
 Component & (val) & (test) & (test) & (train) & (train) \\
\midrule
                            $1$ & 0.82 & \textbf{3.10} & \textbf{2.78} & 1.35 & 2.82 \\ 
                            \underline{$\alpha$}  & \textbf{0.79} & 3.14 & 2.80 & \textbf{1.30} & \textbf{2.74}\\
\midrule
                            replace & 0.88 & 3.16 & 2.94 & 1.41 & 2.79 \\ 
                            \underline{concatenate}  & \textbf{0.79} & \textbf{3.14} & \textbf{2.80} & \textbf{1.30} & \textbf{2.74}\\
\midrule
                            w/o residual & 0.88 & \textbf{3.13} & 2.83 & 1.40 & 2.75 \\ 
                            \underline{w/ residual}  & \textbf{0.79} & 3.14 & \textbf{2.80} & \textbf{1.30} & \textbf{2.74}\\
\bottomrule
\end{tabularx}
\vspace{0pt} 
\caption{\textbf{Ablation experiment results.} Settings used in our final model are underlined.}
\label{Tab:Ablations}
\end{table}
To verify our design, we conducted the following ablation experiments. 
We first compare the performance of the tested variants of the model, where positional attention replaces (p only) or adds 
to (+p) the self-similarity attention, as presented in Table~\ref{Tab:Results}. 
We find that self-similarity is sufficient to achieve the 
performance improvements, with the positional encoding only helping for the KITTI dataset. 
This coincides with our intuition that long-range connections are helpful and that distance-based 
suppression is unnecessary.
In addition, we ablate over
three design choices: (1) learning the  scalar parameter $\alpha$ vs fixing it at $1$, (2) concatenating with 
local motion features vs replacing local motion features, and (3) using a residual connection (adding the output of 
the aggregator to the local motion features) vs not using residual connection (directly concatenating the output of 
the aggregator with the motion features and context features). The results are shown in Table~\ref{Tab:Ablations}.

The key experiment here is showing that concatenation is an important part of the network design. The hypothesis was that 
the network should learn how to select or combine the local and globally-aggregated features, based on some implicit measure
of uncertainty. That is, it is not helpful to replace local features in most non-occluded regions, where they may be more 
reliable and precise than the aggregated features. While the residual connection may also be able to handle this, using both
mechanisms leads to the best performance.

\subsection{Timing, Parameter Counts and Memory}
\label{Sec:overhead}
\setlength\tabcolsep{8pt}
\begin{table}[t!]
\centering
\newcolumntype{C}{>{\centering\arraybackslash}X}
\begin{tabularx}{\columnwidth}{l C C}
\toprule
 Metric & RAFT [38] & Ours \\
\midrule
                            Parameters & 5.3M & 5.9M  \\ 
                            Timing & 60ms & 72ms \\ 
                            GPU Memory & 16.0GB & 17.7GB \\
\bottomrule
\end{tabularx}
\vspace{0pt} 
\caption{\textbf{Timing, parameters and memory.} The GMA module has a modest computational overhead.}
\label{Tab:Computational Overhead}
\end{table}
We demonstrate that the computational overhead of GMA is low relative to the performance improvement, as shown in
Table~\ref{Tab:Computational Overhead}.
The parameter count for our model is 5.9M compared to RAFT which is 5.3M.
We tested the inference time on a single RTX 3090 GPU, with RAFT taking 60ms on average and ours taking 72ms for a single pair of image in the Sintel dataset.
The image size is 436$\times$1024.
The GRU iteration number is set to 12. 
We also tested the GPU memory consumption for training. When training on FlyingChairs on a single 3090 card,
with a random crop of 368$\times$496 and batch size of 8, RAFT takes 16.0GB memory while our network takes
17.2GB memory. We can see that overall the computational overhead is modest while the improvement in results
is significant.

\section{Discussion}
\label{Sec:discussion}

We have demonstrated empirically that long-range connections, weighted by image self-similarities, are very effective at resolving the optical flow of occluded 3D points.
The intuition is that if the network can determine which non-occluded points are moving in the same way, this information can be transmitted to `in-paint' the motion of the occluded points.
Determining which points have similar motion characteristics is a non-trivial task and relies on
the exploitation of statistical biases. 
Similar flow vectors are frequently observed for points belonging to the same
class, due to the homogeneous motion in 3D.
This suggests that we should enable the network to aggregate over 
motions of the same scene objects, 
which motivates our choice to explicitly expose the self-similarity of image features to our GMA module. 
However, additive aggregation of this kind is only helpful when the flow field of the attended locations is approximately homogeneous.
This does not hold exactly for general object and camera motions, where the flow fields may be far from homogeneous, even on the same rigid object.
An example is an object that is directly in front of the camera and rotating about the optical axis,
where the flow vectors are in opposite directions. 
To deal with such scenarios, one possible future work is to first transform the motion features
based on the relative positions and perform aggregation afterwards. 

\section{Conclusion}
Occlusions have long been considered a significant challenge and a major source of error in optical flow estimation. 
Inspired by the recent success of transformers, we introduce a global motion aggregation module to globally
aggregate motion features based on appearance self-similarity of the first image. This has been validated
by experiments that show significantly improved optical flow predictions for occluded regions, particularly the
large reduction of EPE on Sintel Clean and Final.
Our approach of aggregating information over long-range connections using self-similarity is a
simple and effective way to introduce higher-order reasoning into the optical flow problem and
is applicable to any supervised flow network. We expect that further development of aggregation mechanisms
or alternatives would lead to additional performance improvements.

\section*{Acknowledgements}
This research is funded in part by the ARC Centre of Excellence for Robotic Vision 
(CE140100016), ARC Discovery Project grant (DP200102274) and (DP190102261), and 
Continental AG (DFR02541, D.C.). 
S.J. would like to thank Jing Zhang and Yujiao Shi for helpful discussions. 
We thank the anonymous reviewers for their valuable comments.

\part*{Appendix}

\section{Screenshots of Sintel Server Results}
\begin{figure*}[t!]
     \centering
     \begin{subfigure}[b]{\textwidth}
         \centering
         \includegraphics[width=\textwidth]{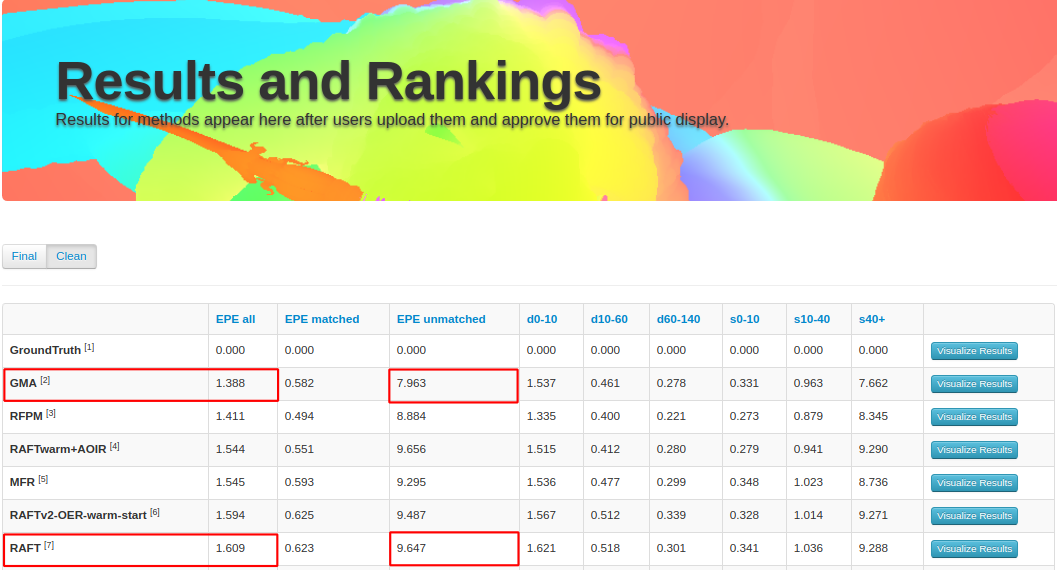}
         \caption{Screenshot for Sintel Clean results.}
     \end{subfigure}%
     \hfill
     \begin{subfigure}[b]{\textwidth}
         \centering
         \includegraphics[width=\textwidth]{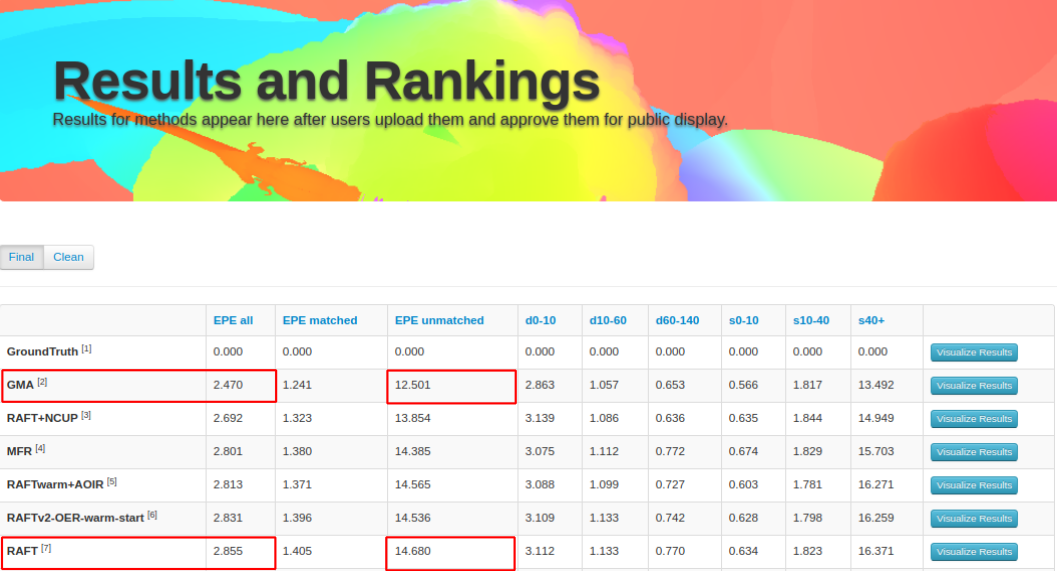}
         \caption{Screenshot for Sintel Final results.}
     \end{subfigure}%
     \caption{\textbf{Screenshots for Sintel Clean and Final results on the test server.} Our
     proposed approach GMA ranks first on both datasets under the `EPE all' metric as of
     March 17th, 2021. We also
     rank first under the `EPE unmatched' metric, with a large margin over previous 
     approaches. This signifies the usefulness of addressing the occlusion problem in optical
     flow.}
     \label{Fig:Screenshots}
\end{figure*}

The screenshots for Sintel Clean and Final results on the test server are shown in 
Figure~\ref{Fig:Screenshots}. We have obtained the best overall results under the
`EPE all' metric. We have also obtained the best results under the `EPE unmatched' metric
with a large margin over previous approaches, which signifies the effectiveness of our
approach in addressing the occlusion problem in optical flow estimation.

\section{Additional Qualitative Results}
\begin{figure*}[ht!]
     \centering
     \begin{subfigure}[b]{0.195\textwidth}
         \centering
         \caption*{Frame 1}
         \includegraphics[width=\textwidth]{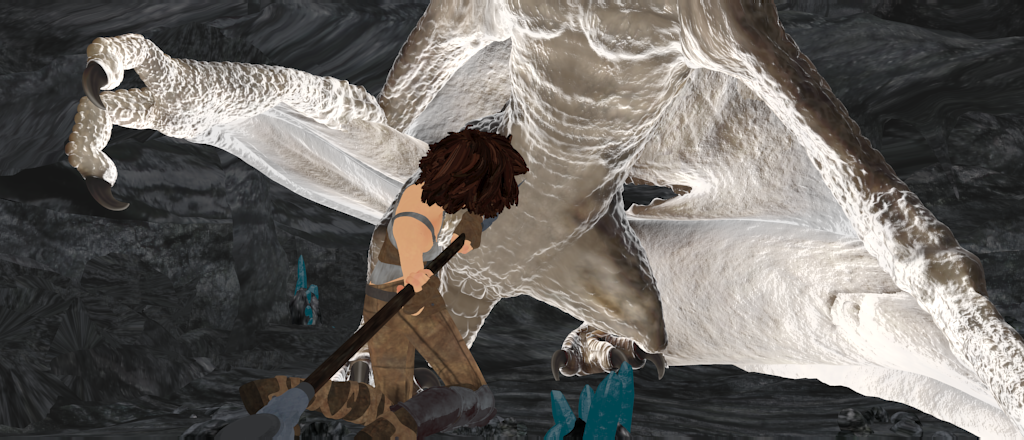}
     \end{subfigure}%
     \hfill
     \begin{subfigure}[b]{0.195\textwidth}
         \centering
         \caption*{Frame 2}
         \includegraphics[width=\textwidth]{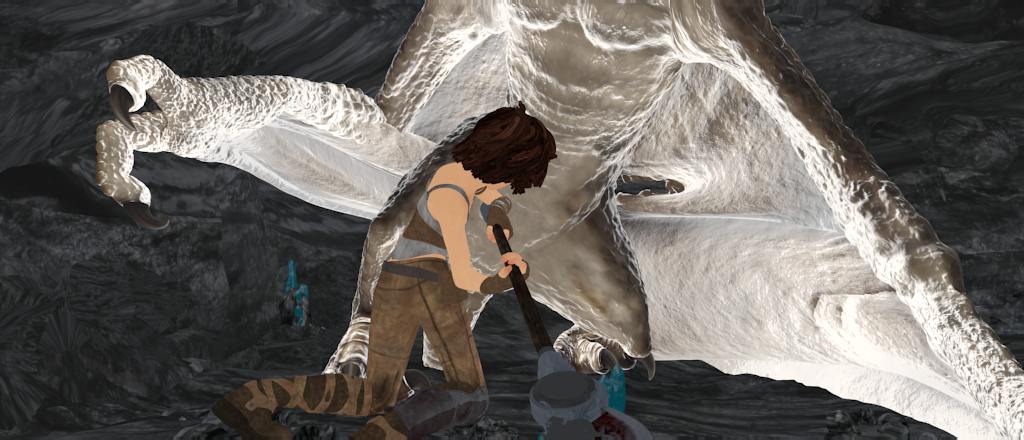}
     \end{subfigure}%
     \hfill
     \begin{subfigure}[b]{0.195\textwidth}
         \centering
         \caption*{Ground-truth}
         \includegraphics[width=\textwidth]{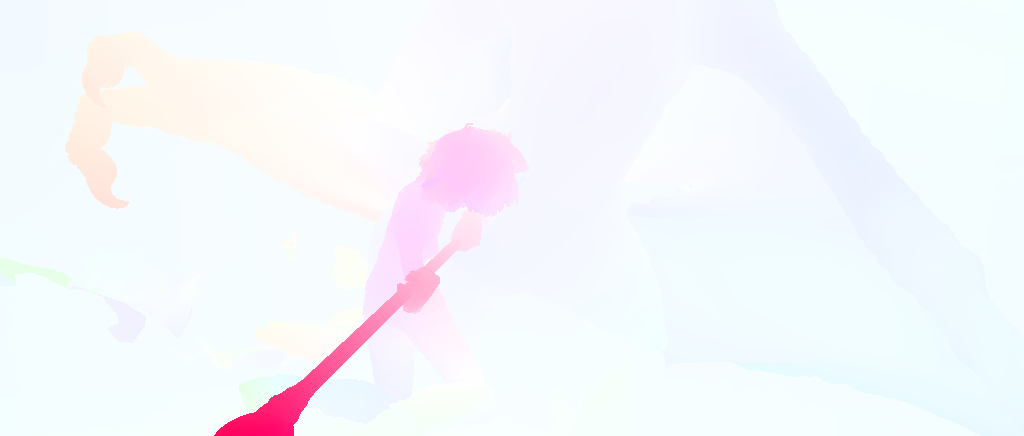}
     \end{subfigure}%
     \hfill
     \begin{subfigure}[b]{0.195\textwidth}
         \centering
         \caption*{RAFT [38]}
         \includegraphics[width=\textwidth]{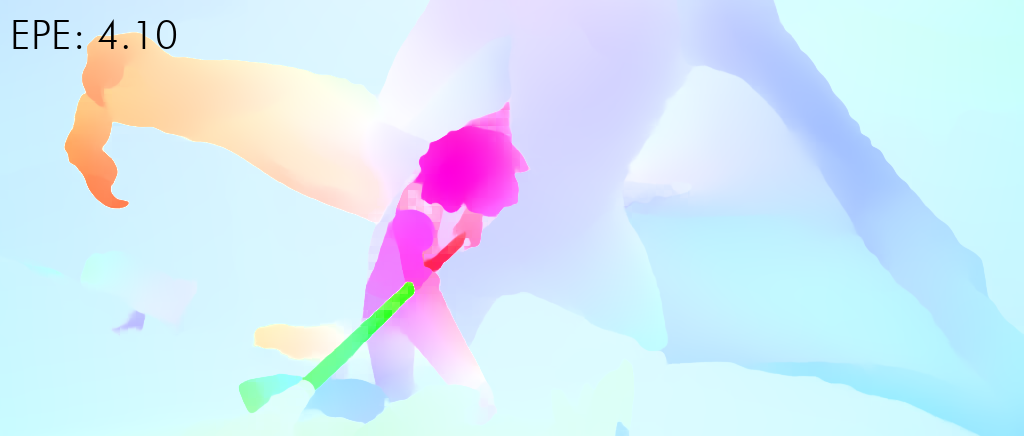}
     \end{subfigure}%
     \hfill
     \begin{subfigure}[b]{0.195\textwidth}
         \centering
         \caption*{Ours}
         \includegraphics[width=\textwidth]{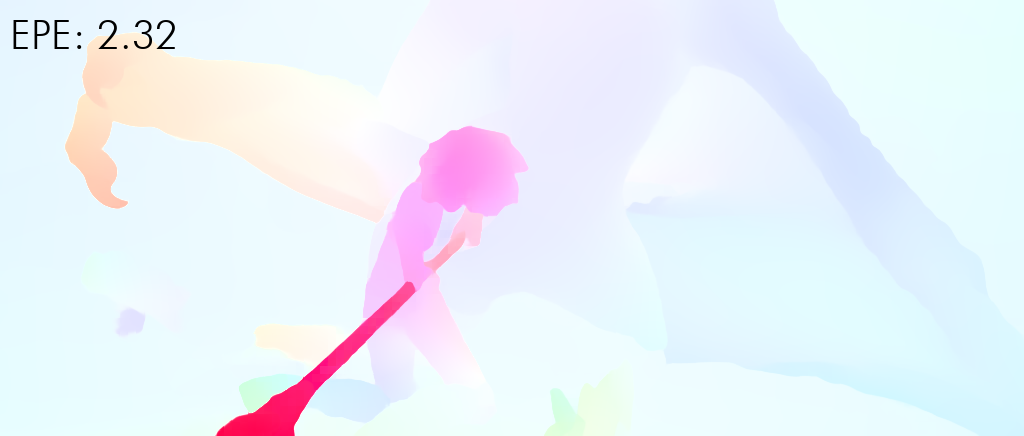}
     \end{subfigure}%
     
     \begin{subfigure}[b]{0.195\textwidth}
         \centering
         \includegraphics[width=\textwidth]{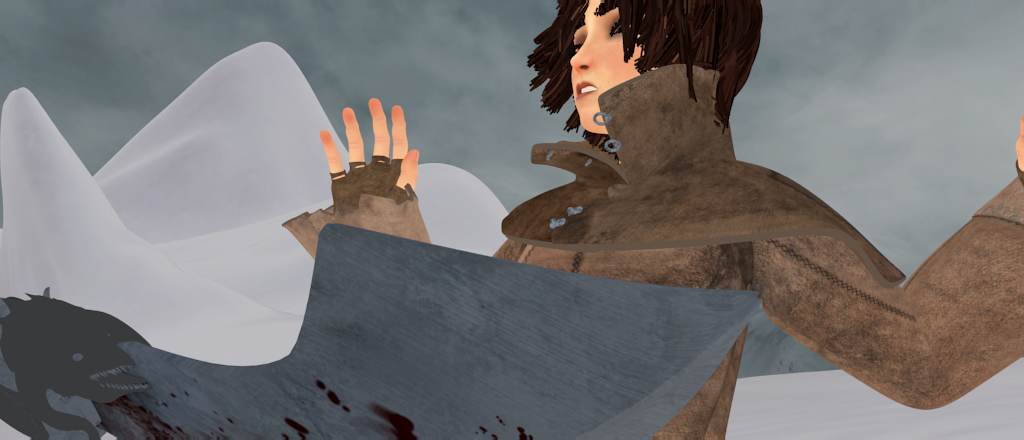}
     \end{subfigure}%
     \hfill
     \begin{subfigure}[b]{0.195\textwidth}
         \centering
         \includegraphics[width=\textwidth]{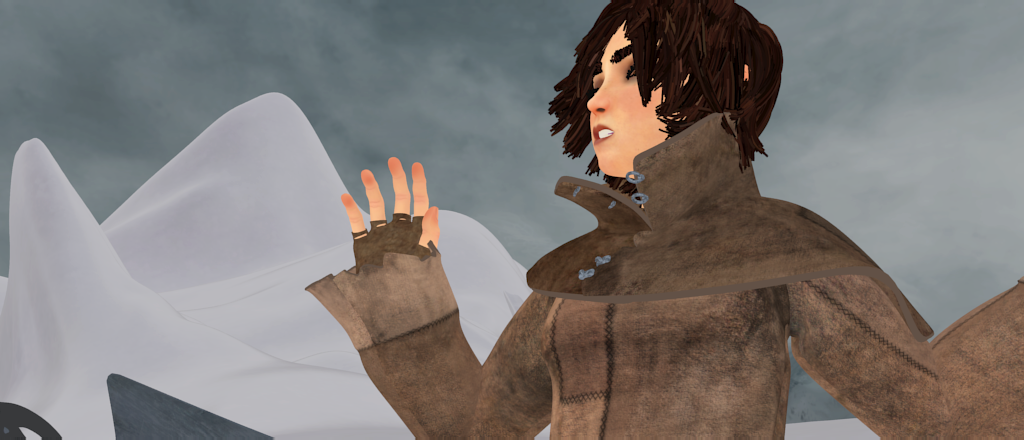}
     \end{subfigure}%
     \hfill
     \begin{subfigure}[b]{0.195\textwidth}
         \centering
         \includegraphics[width=\textwidth]{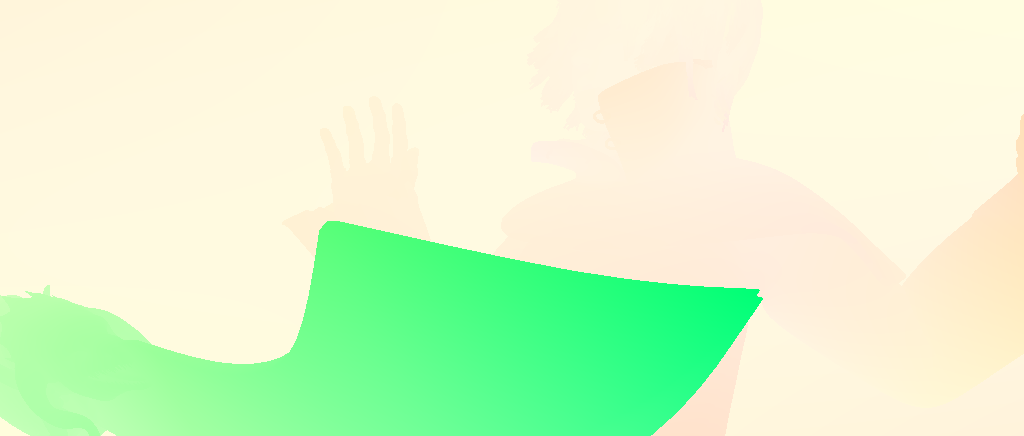}
     \end{subfigure}%
     \hfill
     \begin{subfigure}[b]{0.195\textwidth}
         \centering
         \includegraphics[width=\textwidth]{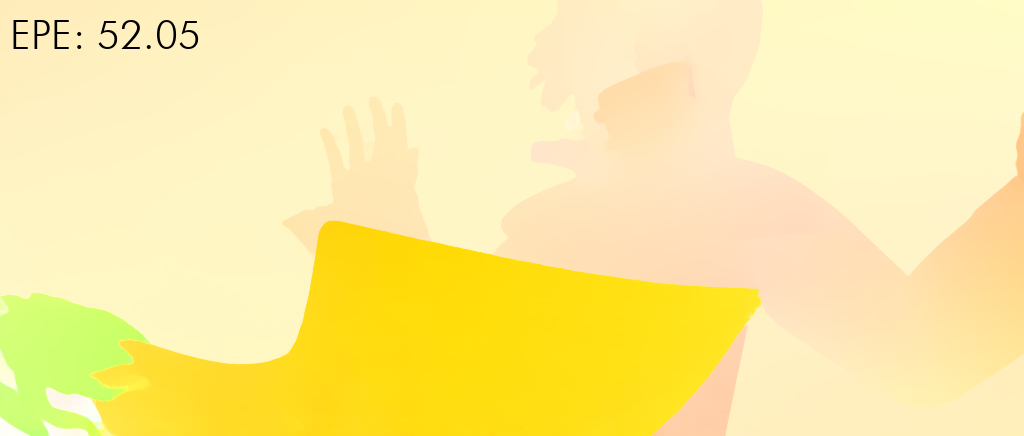}
     \end{subfigure}%
     \hfill
     \begin{subfigure}[b]{0.195\textwidth}
         \centering
         \includegraphics[width=\textwidth]{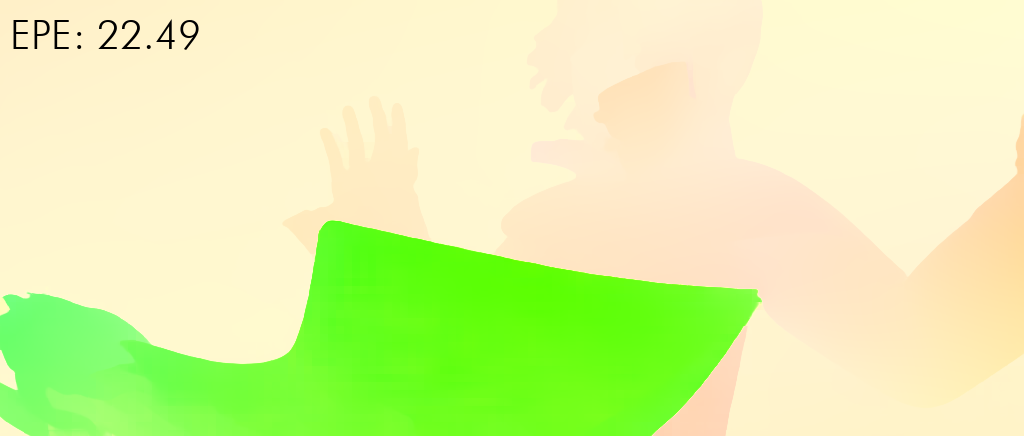}
     \end{subfigure}%
     
    \caption{\textbf{Additional visualisations evaluated on the Sintel Albedo training dataset.} }
    \label{Fig:vis_albedo}
\end{figure*}

Additional visualisations evaluated on the Sintel Albedo training dataset are shown in Figure~\ref{Fig:vis_albedo}.
Note that training has not been conducted on this dataset. 

\begin{figure*}[ht!]
     \centering
     \begin{subfigure}[b]{0.245\textwidth}
         \centering
         \caption*{Frame 1}
         \includegraphics[width=\textwidth]{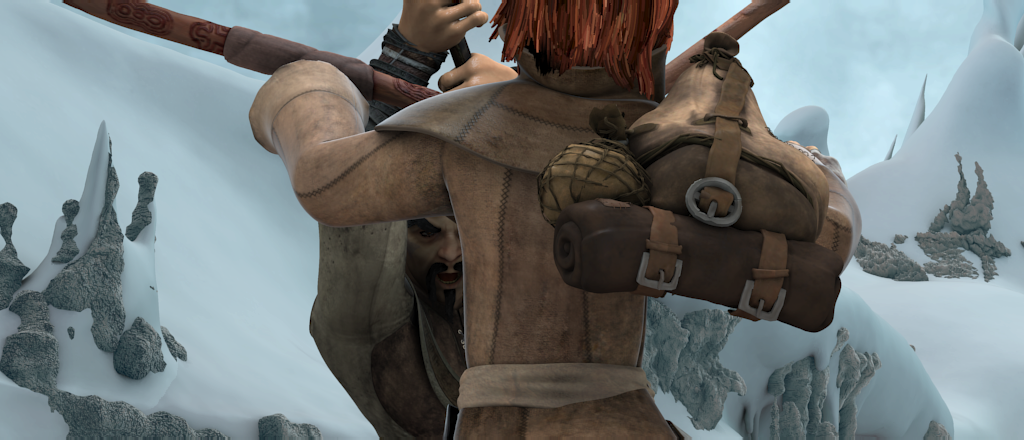}
     \end{subfigure}%
     \hfill
     \begin{subfigure}[b]{0.245\textwidth}
         \centering
         \caption*{Frame 2}
         \includegraphics[width=\textwidth]{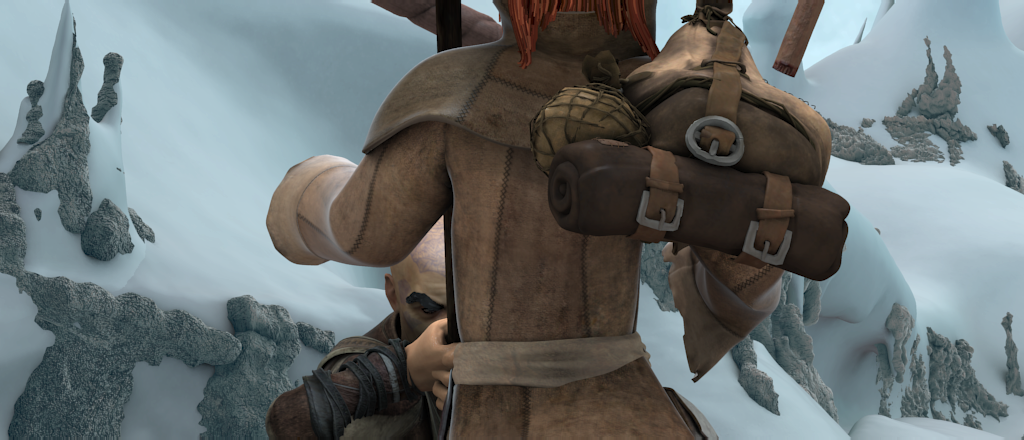}
     \end{subfigure}%
     \hfill
     \begin{subfigure}[b]{0.245\textwidth}
         \centering
         \caption*{RAFT [38]}
         \includegraphics[width=\textwidth]{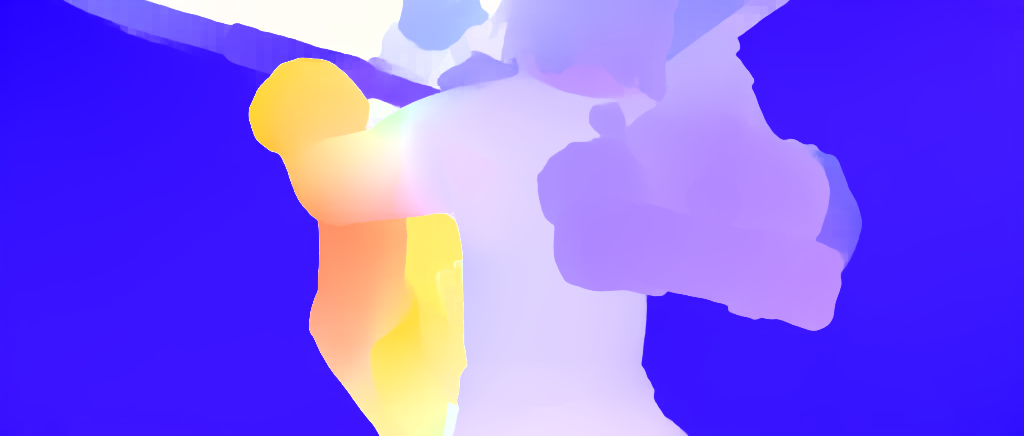}
     \end{subfigure}%
     \hfill
     \begin{subfigure}[b]{0.245\textwidth}
         \centering
         \caption*{Ours}
         \includegraphics[width=\textwidth]{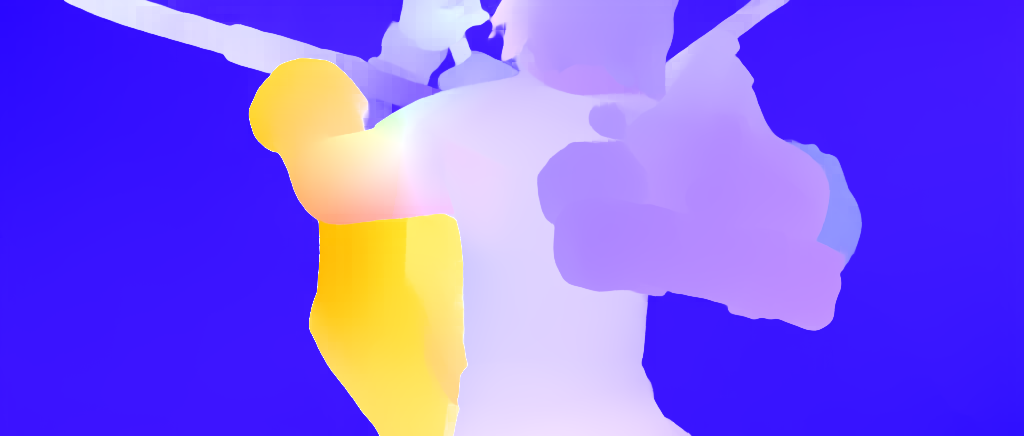}
     \end{subfigure}%
     
     \begin{subfigure}[b]{0.245\textwidth}
         \centering
         \includegraphics[width=\textwidth]{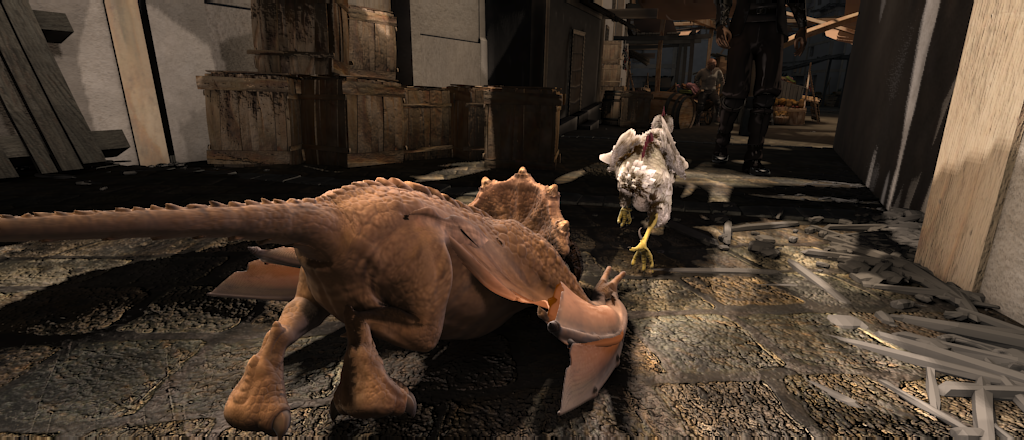}
     \end{subfigure}%
     \hfill
     \begin{subfigure}[b]{0.245\textwidth}
         \centering
         \includegraphics[width=\textwidth]{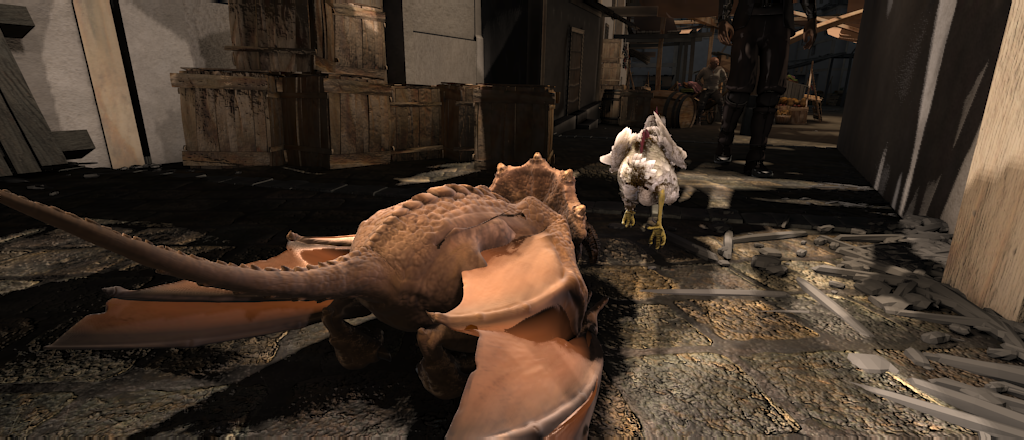}
     \end{subfigure}%
     \hfill
     \begin{subfigure}[b]{0.245\textwidth}
         \centering
         \includegraphics[width=\textwidth]{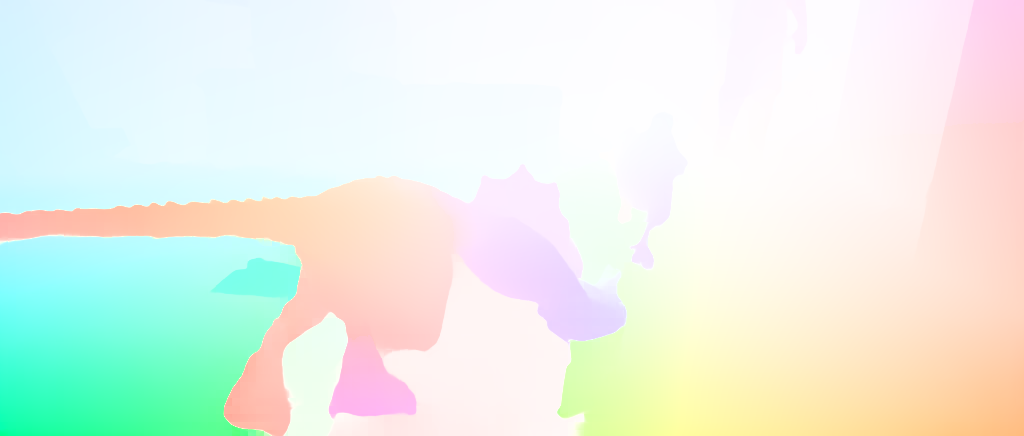}
     \end{subfigure}%
     \hfill
     \begin{subfigure}[b]{0.245\textwidth}
         \centering
         \includegraphics[width=\textwidth]{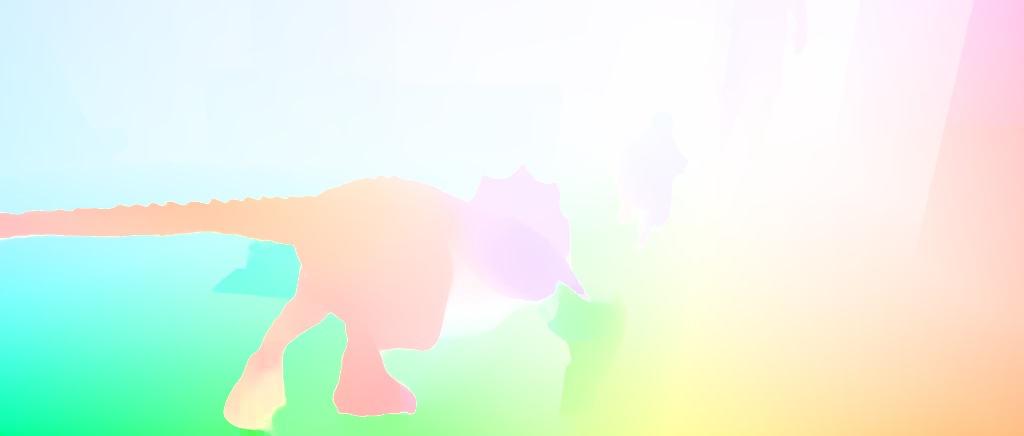}
     \end{subfigure}%
     
     \caption{\textbf{Additional visualisations evaluated on the Sintel Clean test dataset.} }
    \label{Fig:vis_clean}
\end{figure*}

We also give additional visualisations for Sintel Clean and Sintel Final test dataset in
Figure~\ref{Fig:vis_clean} and Figure~\ref{Fig:vis_final} respectively. Since no ground-truth
is provided for Sintel test set, we cannot give an average EPE for each image. 

Additionally, we provide qualitative results on a real-world dataset Slow Flow \cite{janai2017slow}
to demonstrate the benefits of our approach on real-world data. 

\begin{figure*}[ht!]
     \centering
     \begin{subfigure}[b]{0.245\textwidth}
         \centering
         \caption*{Frame 1}
         \includegraphics[width=\textwidth]{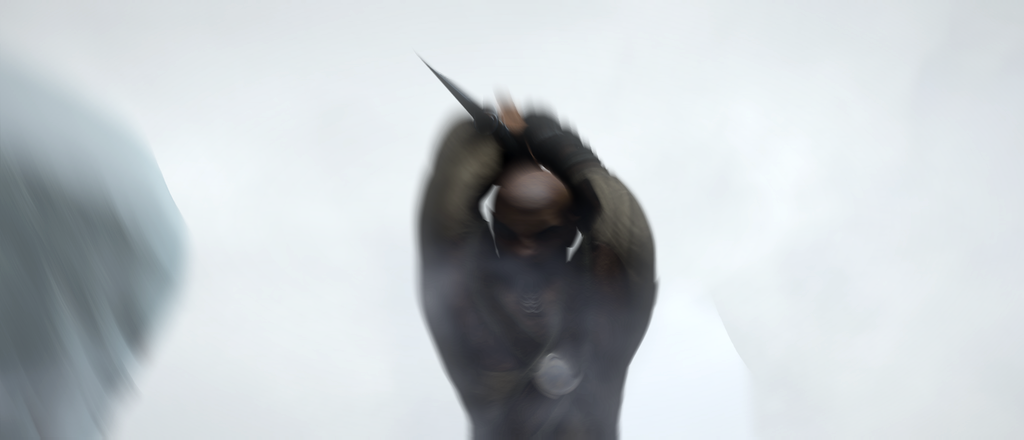}
     \end{subfigure}%
     \hfill
     \begin{subfigure}[b]{0.245\textwidth}
         \centering
         \caption*{Frame 2}
         \includegraphics[width=\textwidth]{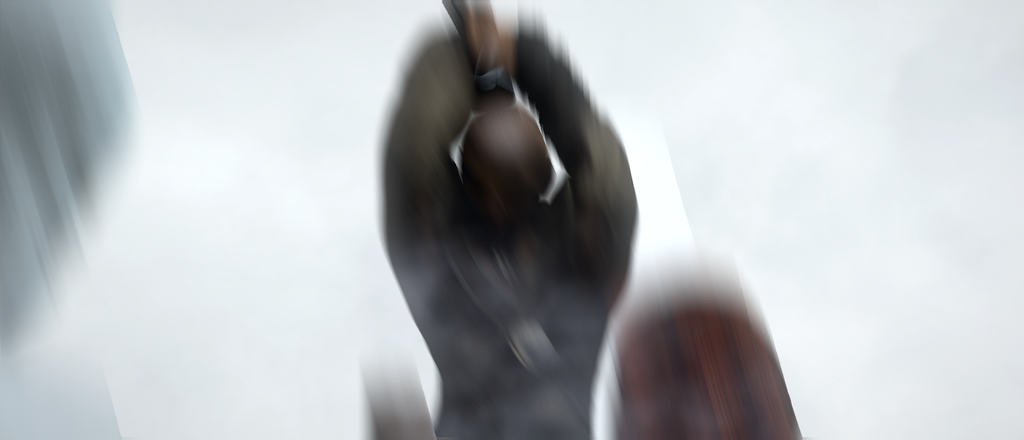}
     \end{subfigure}%
     \hfill
     \begin{subfigure}[b]{0.245\textwidth}
         \centering
         \caption*{RAFT [38]}
         \includegraphics[width=\textwidth]{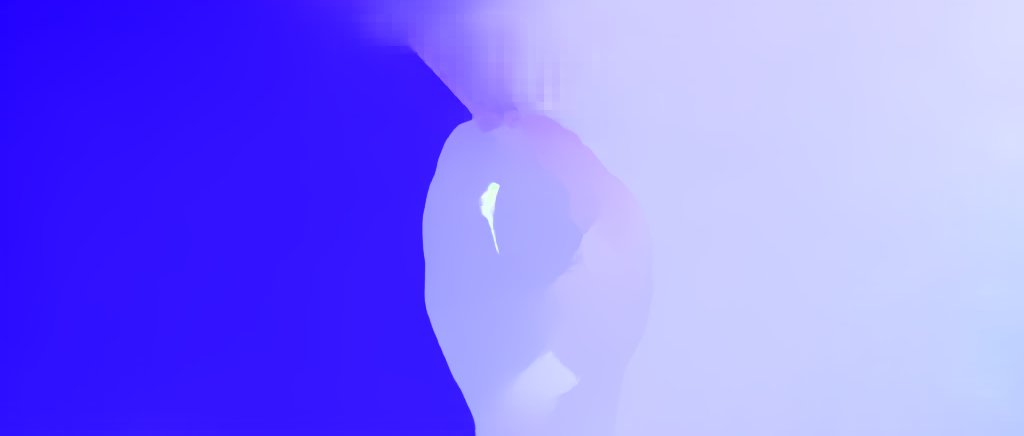}
     \end{subfigure}%
     \hfill
     \begin{subfigure}[b]{0.245\textwidth}
         \centering
         \caption*{Ours}
         \includegraphics[width=\textwidth]{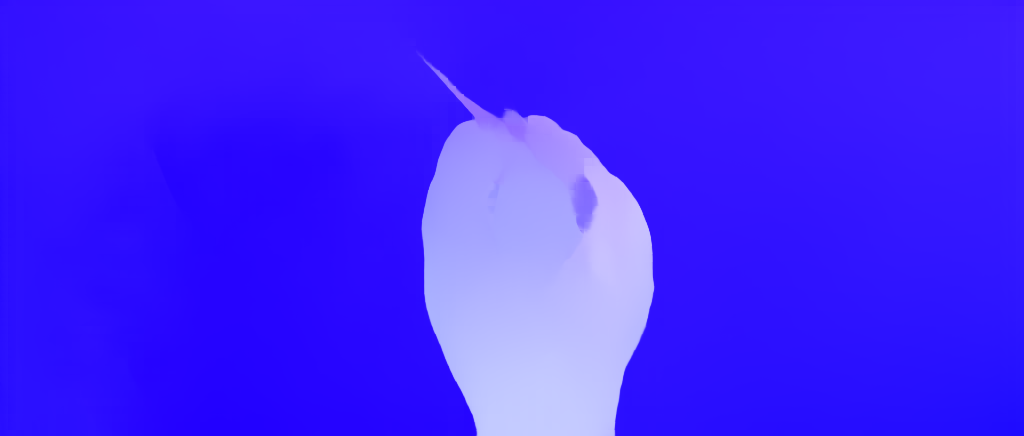}
     \end{subfigure}%
     
     \begin{subfigure}[b]{0.245\textwidth}
         \centering
         \includegraphics[width=\textwidth]{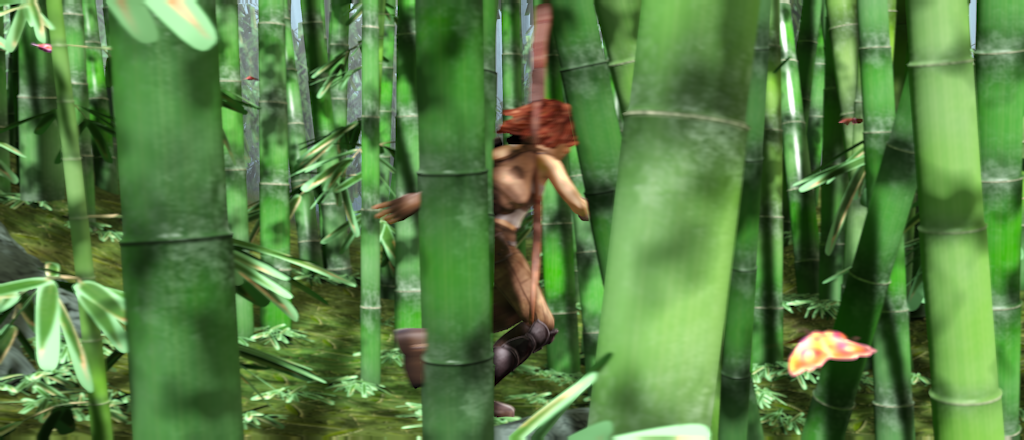}
     \end{subfigure}%
     \hfill
     \begin{subfigure}[b]{0.245\textwidth}
         \centering
         \includegraphics[width=\textwidth]{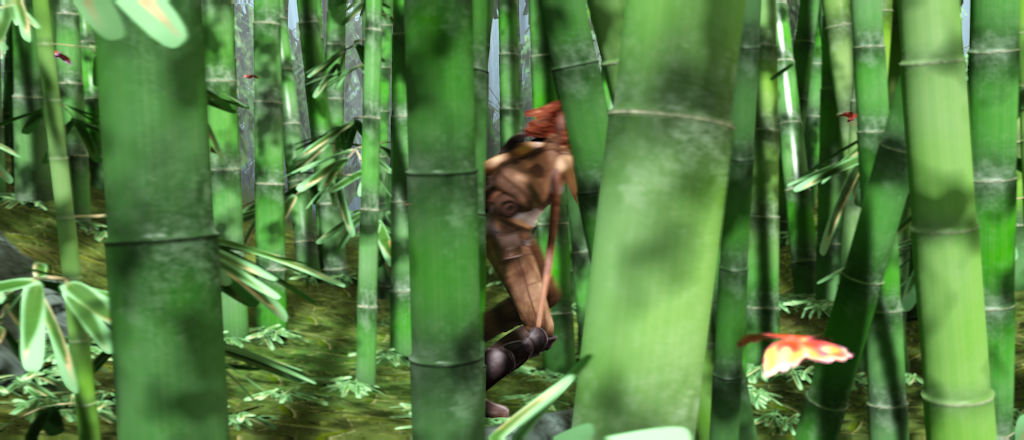}
     \end{subfigure}%
     \hfill
     \begin{subfigure}[b]{0.245\textwidth}
         \centering
         \includegraphics[width=\textwidth]{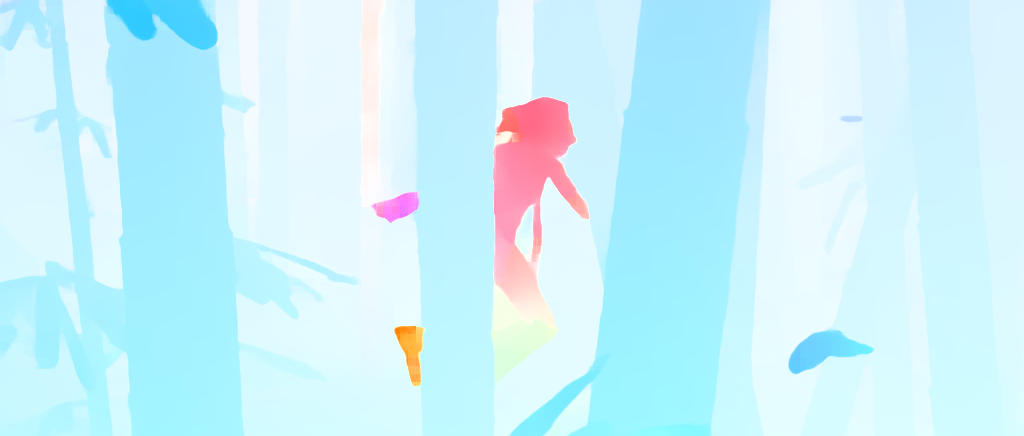}
     \end{subfigure}%
     \hfill
     \begin{subfigure}[b]{0.245\textwidth}
         \centering
         \includegraphics[width=\textwidth]{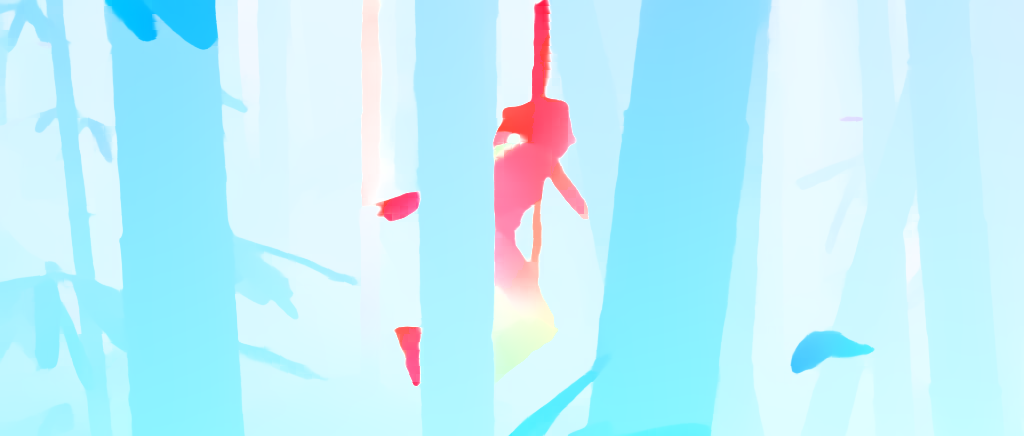}
     \end{subfigure}%
     
     \caption{\textbf{Additional visualisations evaluated on the Sintel Final test dataset.} }
    \label{Fig:vis_final}
\end{figure*}
\begin{figure*}[ht!]
     \centering
     \begin{subfigure}[b]{0.245\textwidth}
         \centering
         \caption*{Frame 1}
         \includegraphics[width=\textwidth]{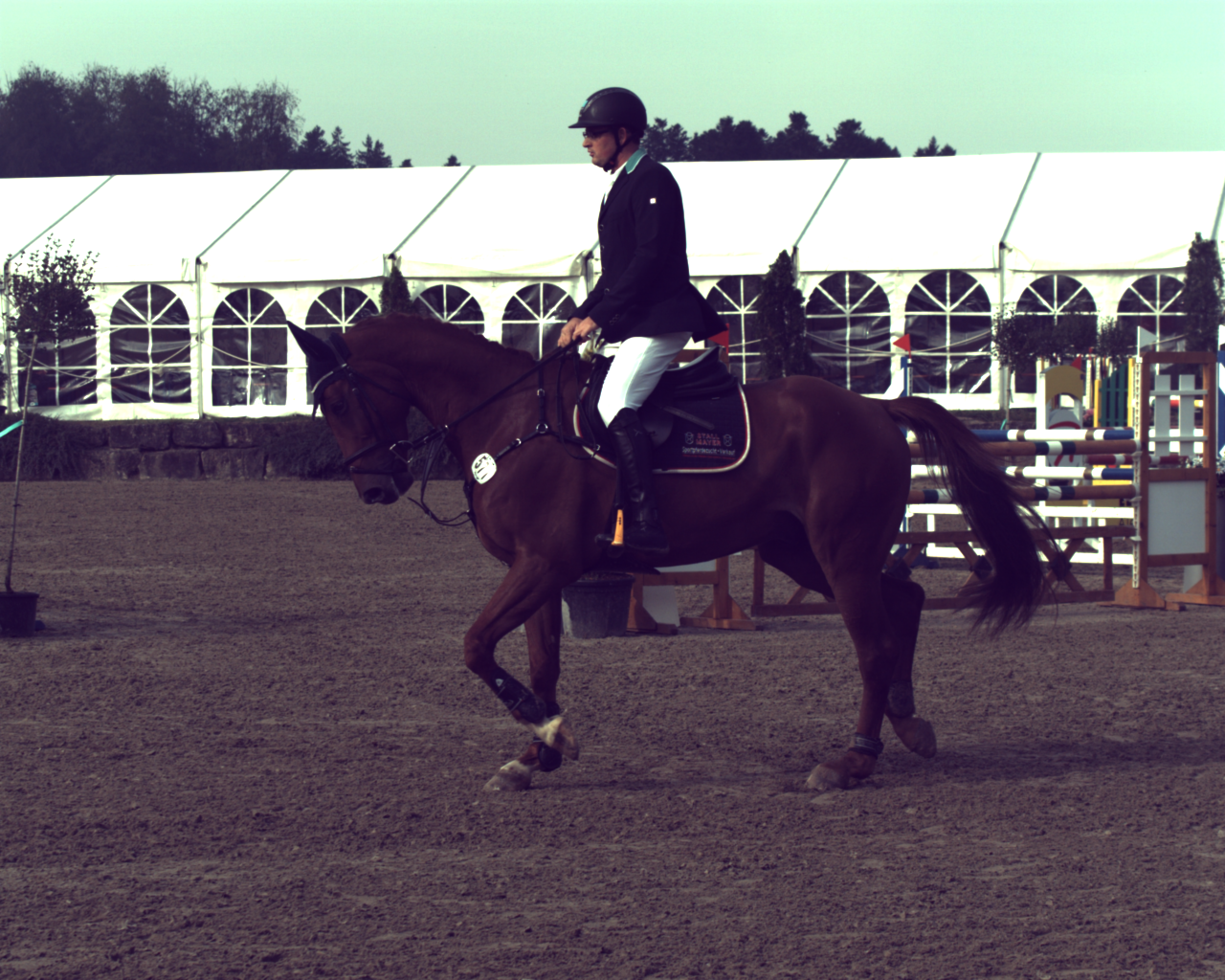}
     \end{subfigure}%
     \hfill
     \begin{subfigure}[b]{0.245\textwidth}
         \centering
         \caption*{Frame 2}
         \includegraphics[width=\textwidth]{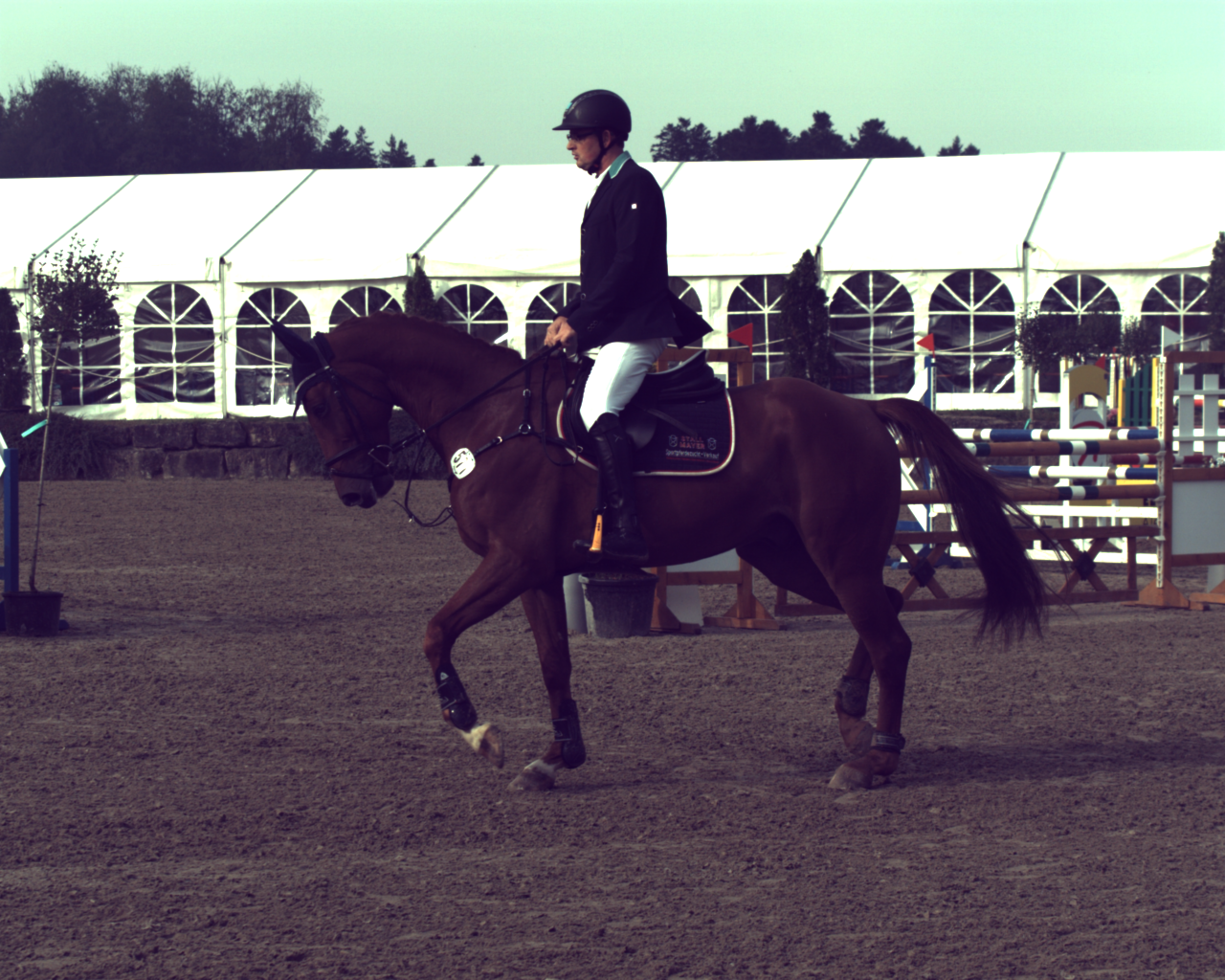}
     \end{subfigure}%
     \hfill
     \begin{subfigure}[b]{0.245\textwidth}
         \centering
         \caption*{RAFT [38]}
         \includegraphics[width=\textwidth]{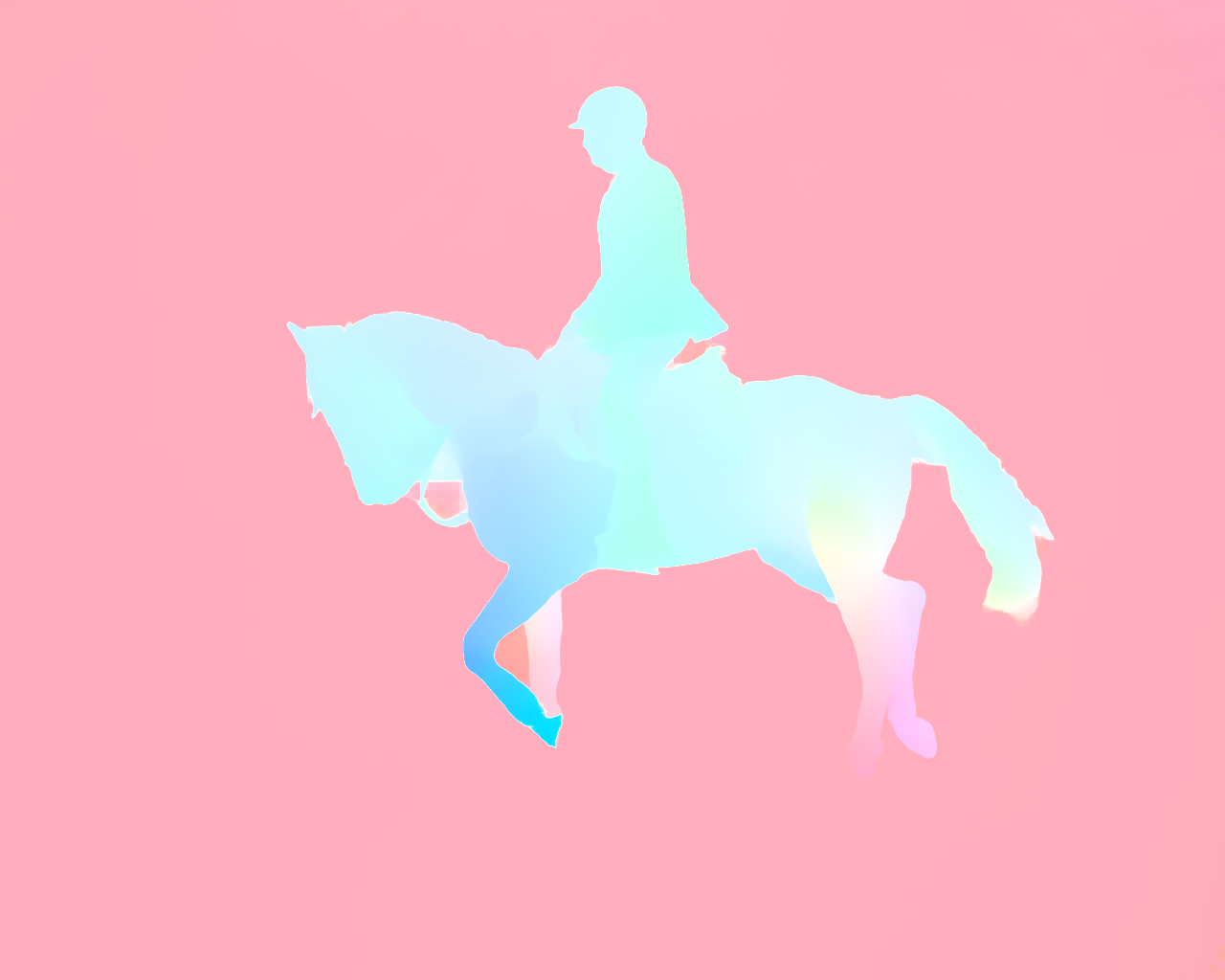}
     \end{subfigure}%
     \hfill
     \begin{subfigure}[b]{0.245\textwidth}
         \centering
         \caption*{Ours}
         \includegraphics[width=\textwidth]{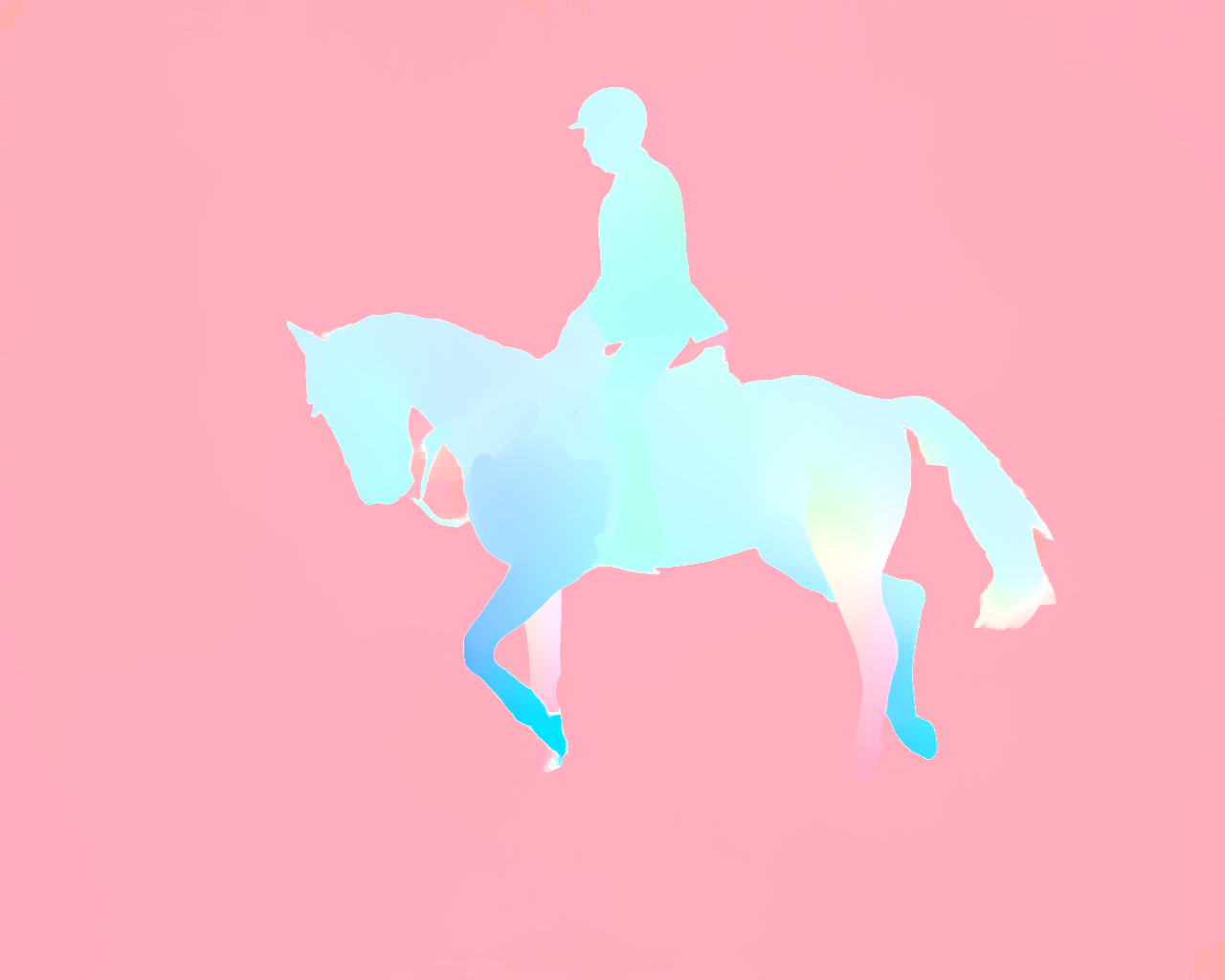}
     \end{subfigure}%
     
     \begin{subfigure}[b]{0.245\textwidth}
         \centering
         \includegraphics[width=\textwidth]{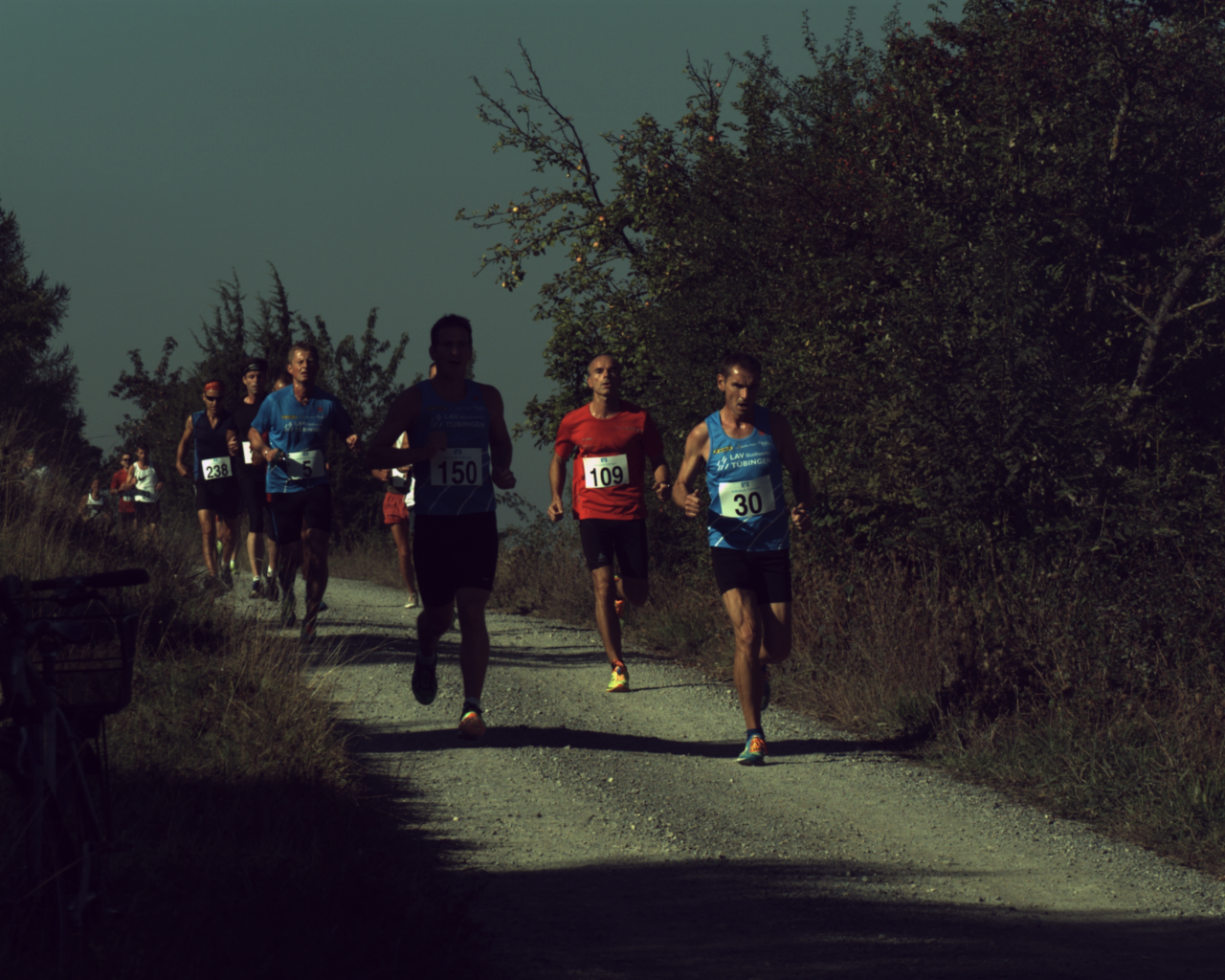}
     \end{subfigure}%
     \hfill
     \begin{subfigure}[b]{0.245\textwidth}
         \centering
         \includegraphics[width=\textwidth]{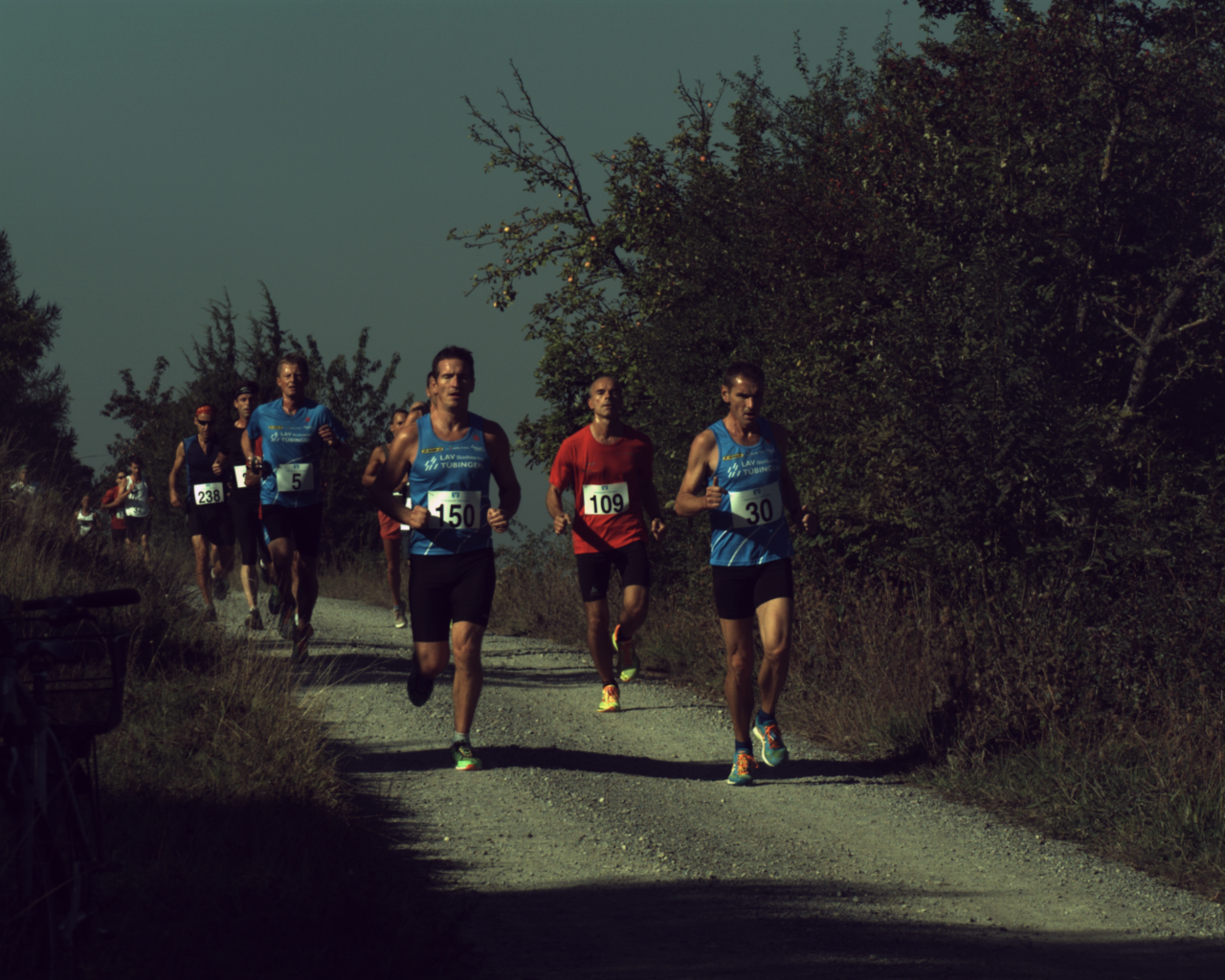}
     \end{subfigure}%
     \hfill
     \begin{subfigure}[b]{0.245\textwidth}
         \centering
         \includegraphics[width=\textwidth]{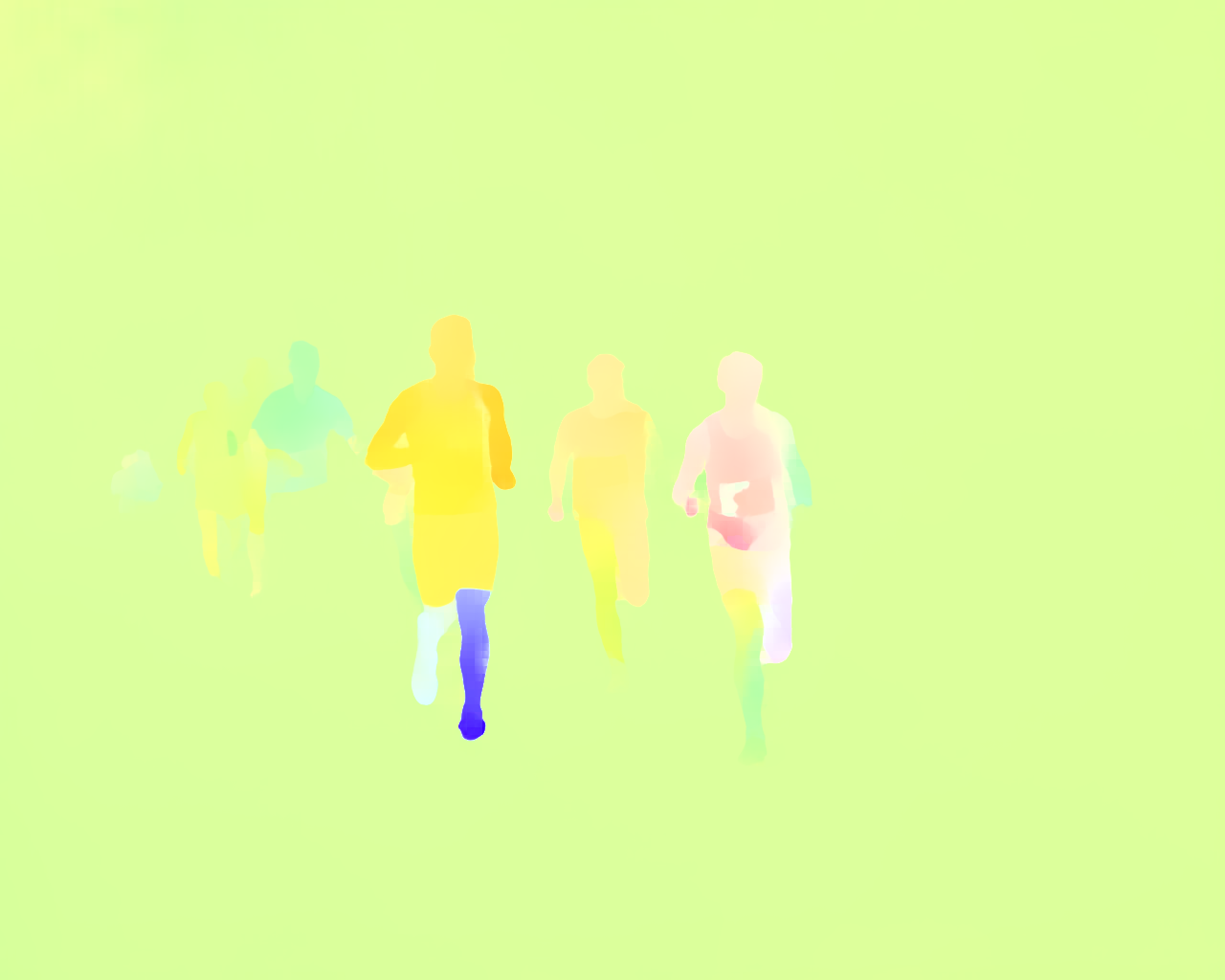}
     \end{subfigure}%
     \hfill
     \begin{subfigure}[b]{0.245\textwidth}
         \centering
         \includegraphics[width=\textwidth]{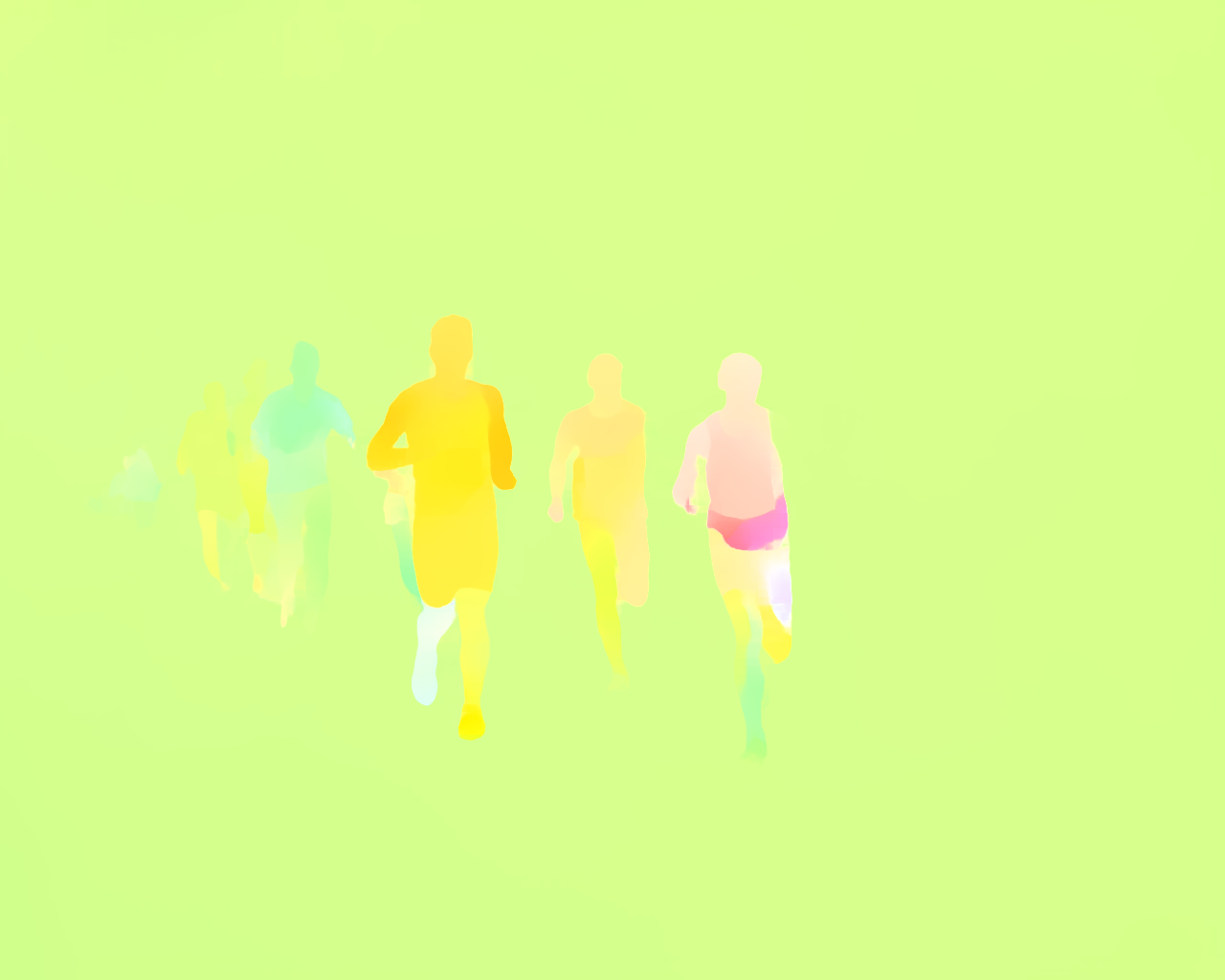}
     \end{subfigure}%
     
     \caption{\textbf{Additional visualisations evaluated on the Sintel Final test dataset.} }
    \label{Fig:vis_final}
\end{figure*}

\clearpage

{\small
\bibliographystyle{ieee_fullname}
\bibliography{egbib}

\begin{thebibliography}{10}\itemsep=-1pt

\bibitem{alvarez}
Luis Alvarez, Rachid Deriche, Th{\'e}o Papadopoulo, and Javier S{\'a}nchez.
\newblock Symmetrical dense optical flow estimation with occlusions detection.
\newblock {\em IJCV}, 2007.

\bibitem{scopeflow}
Aviram Bar-Haim and Lior Wolf.
\newblock Scopeflow: Dynamic scene scoping for optical flow.
\newblock {\em CVPR}, 2020.

\bibitem{bello2019attention}
Irwan Bello, Barret Zoph, Ashish Vaswani, Jonathon Shlens, and Quoc~V Le.
\newblock Attention augmented convolutional networks.
\newblock {\em ICCV}, 2019.

\bibitem{lorentzian}
Michael~J Black and Paul Anandan.
\newblock The robust estimation of multiple motions: Parametric and
  piecewise-smooth flow fields.
\newblock {\em CVIU}, 1996.

\bibitem{boykov}
Yuri Boykov and Vladimir Kolmogorov.
\newblock An experimental comparison of min-cut/max-flow algorithms for energy
  minimization in vision.
\newblock {\em TPAMI}, 2004.

\bibitem{broxl1}
Thomas Brox, Andr{\'e}s Bruhn, Nils Papenberg, and Joachim Weickert.
\newblock High accuracy optical flow estimation based on a theory for warping.
\newblock {\em ECCV}, 2004.

\bibitem{LKmeetsHS}
Andr{\'e}s Bruhn, Joachim Weickert, and Christoph Schn{\"o}rr.
\newblock Lucas/kanade meets horn/schunck: Combining local and global optic
  flow methods.
\newblock {\em IJCV}, 2005.

\bibitem{sintel}
D.~J. Butler, J. Wulff, G.~B. Stanley, and M.~J. Black.
\newblock A naturalistic open source movie for optical flow evaluation.
\newblock {\em ECCV}, 2012.

\bibitem{fullflow}
Qifeng Chen and Vladlen Koltun.
\newblock Full flow: Optical flow estimation by global optimization over
  regular grids.
\newblock {\em CVPR}, 2016.

\bibitem{vit}
Alexey Dosovitskiy, Lucas Beyer, Alexander Kolesnikov, Dirk Weissenborn,
  Xiaohua Zhai, Thomas Unterthiner, Mostafa Dehghani, Matthias Minderer, Georg
  Heigold, Sylvain Gelly, et~al.
\newblock An image is worth 16x16 words: Transformers for image recognition at
  scale.
\newblock {\em ICLR}, 2020.

\bibitem{flownet}
Alexey Dosovitskiy, Philipp Fischer, Eddy Ilg, Philip Hausser, Caner Hazirbas,
  Vladimir Golkov, Patrick Van Der~Smagt, Daniel Cremers, and Thomas Brox.
\newblock Flownet: Learning optical flow with convolutional networks.
\newblock {\em ICCV}, 2015.

\bibitem{dual}
Jun Fu, Jing Liu, Haijie Tian, Yong Li, Yongjun Bao, Zhiwei Fang, and Hanqing
  Lu.
\newblock Dual attention network for scene segmentation.
\newblock {\em CVPR}, 2019.

\bibitem{hinton1976using}
G Hinton.
\newblock Using relaxation to find a puppet.
\newblock {\em Proceedings of the 2nd Summer Conference on Artificial
  Intelligence and Simulation of Behaviour}, 1976.

\bibitem{horn}
Berthold~KP Horn and Brian~G Schunck.
\newblock Determining optical flow.
\newblock {\em Techniques and Applications of Image Understanding}, 1981.

\bibitem{hosni2012fast}
Asmaa Hosni, Christoph Rhemann, Michael Bleyer, Carsten Rother, and Margrit
  Gelautz.
\newblock Fast cost-volume filtering for visual correspondence and beyond.
\newblock {\em TPAMI}, 2012.

\bibitem{crisscross}
Zilong Huang, Xinggang Wang, Lichao Huang, Chang Huang, Yunchao Wei, and Wenyu
  Liu.
\newblock Ccnet: Criss-cross attention for semantic segmentation.
\newblock {\em ICCV}, 2019.

\bibitem{liteflownet}
Tak-Wai Hui, Xiaoou Tang, and Chen Change~Loy.
\newblock Liteflownet: A lightweight convolutional neural network for optical
  flow estimation.
\newblock {\em CVPR}, 2018.

\bibitem{mirrorflow}
Junhwa Hur and Stefan Roth.
\newblock Mirrorflow: Exploiting symmetries in joint optical flow and occlusion
  estimation.
\newblock {\em ICCV}, 2017.

\bibitem{irr}
Junhwa Hur and Stefan Roth.
\newblock Iterative residual refinement for joint optical flow and occlusion
  estimation.
\newblock {\em CVPR}, 2019.

\bibitem{flownet2}
Eddy Ilg, Nikolaus Mayer, Tonmoy Saikia, Margret Keuper, Alexey Dosovitskiy,
  and Thomas Brox.
\newblock Flownet 2.0: Evolution of optical flow estimation with deep networks.
\newblock {\em CVPR}, 2017.

\bibitem{multiframe}
Joel Janai, Fatma Guney, Anurag Ranjan, Michael Black, and Andreas Geiger.
\newblock Unsupervised learning of multi-frame optical flow with occlusions.
\newblock {\em ECCV}, 2018.

\bibitem{janai2017slow}
Joel Janai, Fatma Guney, Jonas Wulff, Michael~J Black, and Andreas Geiger.
\newblock Slow flow: Exploiting high-speed cameras for accurate and diverse
  optical flow reference data.
\newblock {\em CVPR}, 2017.

\bibitem{scv}
Shihao Jiang, Yao Lu, Hongdong Li, and Richard Hartley.
\newblock Learning optical flow from a few matches.
\newblock {\em CVPR}, 2021.

\bibitem{tracking}
Dieter Koller, Joseph Weber, and Jitendra Malik.
\newblock Robust multiple car tracking with occlusion reasoning.
\newblock {\em ECCV}, 1994.

\bibitem{hd1k}
Daniel Kondermann, Rahul Nair, Katrin Honauer, Karsten Krispin, Jonas Andrulis,
  Alexander Brock, Burkhard Gussefeld, Mohsen Rahimimoghaddam, Sabine Hofmann,
  Claus Brenner, et~al.
\newblock The hci benchmark suite: Stereo and flow ground truth with
  uncertainties for urban autonomous driving.
\newblock {\em CVPR Workshop}, 2016.

\bibitem{selflow}
Pengpeng Liu, Michael Lyu, Irwin King, and Jia Xu.
\newblock Selflow: Self-supervised learning of optical flow.
\newblock {\em CVPR}, 2019.

\bibitem{things}
N. Mayer, E. Ilg, P. H{\"a}usser, P. Fischer, D. Cremers, A. Dosovitskiy, and
  T. Brox.
\newblock A large dataset to train convolutional networks for disparity,
  optical flow, and scene flow estimation.
\newblock {\em CVPR}, 2016.

\bibitem{discreteflow}
Moritz Menze, Christian Heipke, and Andreas Geiger.
\newblock Discrete optimization for optical flow.
\newblock {\em GCPR}, 2015.

\bibitem{kitti}
Moritz Menze, Christian Heipke, and Andreas Geiger.
\newblock Joint 3d estimation of vehicles and scene flow.
\newblock {\em ISPRS Workshop on Image Sequence Analysis}, 2015.

\bibitem{pytorch}
Adam Paszke, Sam Gross, Soumith Chintala, Gregory Chanan, Edward Yang, Zachary
  DeVito, Zeming Lin, Alban Desmaison, Luca Antiga, and Adam Lerer.
\newblock Automatic differentiation in pytorch.
\newblock {\em NIPS Workshop}, 2017.

\bibitem{standalone}
Prajit Ramachandran, Niki Parmar, Ashish Vaswani, Irwan Bello, Anselm Levskaya,
  and Jonathon Shlens.
\newblock Stand-alone self-attention in vision models.
\newblock {\em NeurIPS}, 2019.

\bibitem{epicflow}
Jerome Revaud, Philippe Weinzaepfel, Zaid Harchaoui, and Cordelia Schmid.
\newblock Epicflow: Edge-preserving interpolation of correspondences for
  optical flow.
\newblock {\em CVPR}, 2015.

\bibitem{onecycle}
Leslie~N Smith and Nicholay Topin.
\newblock Super-convergence: Very fast training of neural networks using large
  learning rates.
\newblock {\em Artificial Intelligence and Machine Learning for Multi-Domain
  Operations Applications}, 2019.

\bibitem{sun2014local}
Deqing Sun, Ce Liu, and Hanspeter Pfister.
\newblock Local layering for joint motion estimation and occlusion detection.
\newblock {\em CVPR}, 2014.

\bibitem{sun2010secrets}
Deqing Sun, Stefan Roth, and Michael~J Black.
\newblock Secrets of optical flow estimation and their principles.
\newblock 2010.

\bibitem{sun2010layered}
Deqing Sun, Erik~B Sudderth, and Michael~J Black.
\newblock Layered image motion with explicit occlusions, temporal consistency,
  and depth ordering.
\newblock {\em NIPS}, 2010.

\bibitem{pwcnet}
Deqing Sun, Xiaodong Yang, Ming-Yu Liu, and Jan Kautz.
\newblock Pwc-net: Cnns for optical flow using pyramid, warping, and cost
  volume.
\newblock {\em CVPR}, 2018.

\bibitem{pwcnet+}
Deqing Sun, Xiaodong Yang, Ming-Yu Liu, and Jan Kautz.
\newblock Models matter, so does training: An empirical study of cnns for
  optical flow estimation.
\newblock {\em TPAMI}, 2019.

\bibitem{raft}
Zachary Teed and Jia Deng.
\newblock Raft: Recurrent all-pairs field transforms for optical flow.
\newblock {\em ECCV}, 2020.

\bibitem{transformer}
Ashish Vaswani, Noam Shazeer, Niki Parmar, Jakob Uszkoreit, Llion Jones,
  Aidan~N Gomez, Lukasz Kaiser, and Illia Polosukhin.
\newblock Attention is all you need.
\newblock {\em NIPS}, 2017.

\bibitem{nonlocal}
Xiaolong Wang, Ross Girshick, Abhinav Gupta, and Kaiming He.
\newblock Non-local neural networks.
\newblock {\em CVPR}, 2018.

\bibitem{occaware}
Yang Wang, Yi Yang, Zhenheng Yang, Liang Zhao, Peng Wang, and Wei Xu.
\newblock Occlusion aware unsupervised learning of optical flow.
\newblock {\em CVPR}, 2018.

\bibitem{wedel2009improved}
Andreas Wedel, Thomas Pock, Christopher Zach, Horst Bischof, and Daniel
  Cremers.
\newblock An improved algorithm for tv-l 1 optical flow.
\newblock {\em Statistical and Geometrical Approaches to Visual Motion
  Analysis}, 2009.

\bibitem{dcflow}
Jia Xu, Ren{\'e} Ranftl, and Vladlen Koltun.
\newblock Accurate optical flow via direct cost volume processing.
\newblock {\em CVPR}, 2017.

\bibitem{vcn}
Gengshan Yang and Deva Ramanan.
\newblock Volumetric correspondence networks for optical flow.
\newblock {\em NeurIPS}, 2019.

\bibitem{hd3}
Zhichao Yin, Trevor Darrell, and Fisher Yu.
\newblock Hierarchical discrete distribution decomposition for match density
  estimation.
\newblock {\em CVPR}, 2019.

\bibitem{tvl1}
Christopher Zach, Thomas Pock, and Horst Bischof.
\newblock A duality based approach for realtime tv-l 1 optical flow.
\newblock {\em Joint Pattern Recognition Symposium}, 2007.

\bibitem{maskflownet}
Shengyu Zhao, Yilun Sheng, Yue Dong, Eric~I Chang, Yan Xu, et~al.
\newblock Maskflownet: Asymmetric feature matching with learnable occlusion
  mask.
\newblock {\em CVPR}, 2020.

\end{thebibliography}
}

\end{document}